# Physics-Informed Bayesian Learning of Electrohydrodynamic Polymer Jet Printing Dynamics

Athanasios Oikonomou[1,4,7], Theodoros Loutas[1], Dixia Fan[2], Alysia Garmulewicz[3], George Nounesis[4], Santanu Chaudhuri[5,6] and Filippos Tourlomousis[4,7,8*]

[1]Mechanical Engineering, University of Patras, Patras, Greece

[2]Westlake University, Hangzhou, China

[3]Faculty of Economics and Administration, University of Santiago, Santiago, Chile

[4]National Centre for Scientific Research Demokritos, Agia Paraskevi, Attica, 1534, Greece

[5]Civil, Materials, and Environmental Engineering Department, University of Illinois at Chicago, IL, 60607, United States

[6]Argonne National Laboratory, Lemont, IL, 60439, United States

[7]Superlabs AMKE, Marousi, Attica, 15124, Greece

[8]Biological Lattice Industries Corp., Boston, MA, 02108, United States

## Abstract

Calibration of highly dynamic multi-physics manufacturing processes such as electro-hydrodynamics-based additive manufacturing (AM) technologies (E-jet printing) is still performed by labor-intensive trial-and-error practices. These practices have hindered the broad adoption of these technologies, demanding a new paradigm of self-calibrating E-jet printing machines. To address this need, we developed GPJet, an end-to-end physics-informed Bayesian learning framework, and tested it on a virtual E-jet printing machine with in-process jet monitoring capabilities. GPJet consists of three modules: a) the Machine Vision module, b) the Physics-Based Modeling Module, and c) the Machine Learning (ML) module. We demonstrate that the Machine Vision module can extract high-fidelity jet features in real-time from video data using an automated parallelized computer vision workflow. In addition, we show that the Machine Vision module, combined with the Physics-based modeling module, can act as closed-loop sensory feedback to the Machine Learning module of high- and low-fidelity data. Powered by our data-centric approach, we demonstrate that the online ML planner can actively learn the jet process dynamics using video and physics with minimum experimental cost. GPJet brings us one step closer to realizing the vision of intelligent AM machines that can efficiently search complex process-structure-property landscapes and create optimized material solutions for a wide range of applications at a fraction of the cost and speed.

# 1. Introduction

The programmable assembly of functional inks in two- and three-dimensions using computer numerically controlled (CNC) machines coupled with printing technologies has revolutionized the design and fabrication of physical objects. Extrusion-based additive manufacturing (AM) technologies, often referred to as direct ink writing or 3D printing, are transforming fields such as healthcare, robotics, electronics, and sustainability [1,2]. While the potential of 3D printing is celebrated very often in scientific journals and the media, there is a "secret" that practitioners and companies of 3D printing do not stress out. This under-reported reality entails the extensive experimentation and manual labor required for tuning process parameters that are high in number and often inter-dependent, to achieve process stability and reproducible outcomes [3]. Every time a new material needs to be processed, or ambient conditions vary, practitioners follow trial and error approaches for printing process calibration. These calibration practices have led to the creation of experienced "super users" at the expense of an enormous degree of individual process engineering.

Electro-hydrodynamics-based AM technologies, also known as E-jet printing technologies, are notable examples of extrusion-based AM technologies that have been facing such challenges due to their complex multi-physics and highly dynamic nature [4,5]. During E-jet printing, a polymer solution or a polymer melt is extruded through a charged needle tip towards a grounded collector. As soon as the electrostatic stresses overcome the polymer material's viscoelastic and surface tension stresses, a cone-jet is formed in the free flow regime (**Figure 1a**). An instabilities area, whose length depends on the nature of the polymer, follows the cone-jet regime. Focusing on the polymer melt case, where the instabilities area is closer to the collector (**Figure 1a**), a translational stage can be employed to write high-resolution fibers (**Figure 1b**), a process known as melt electro-writing (MEW). With this capability, MEW has been established as an emerging high-resolution AM technology for fabricating architected biomaterial scaffolds, opening new tissue engineering avenues. MEW has undergone ten years of process optimization studies since its first inception in the literature [6,7]. Tunable fiber diameter and patterning fidelity are critical scaffold attributes for biological outcomes and efficacy. These can be optimized by tuning five inter-dependent user-controlled process parameters assuming stable ambient environmental conditions (temperature and humidity): a) the applied voltage at the needle tip, b) the extrusion volumetric flowrate, c) the temperature at the syringe, d) the collector speed, and needle tip to collector distance. Considering the dynamic range of each process variable in combination with the highly sensitive spatial and time scales of the process in the micron range, one quickly realizes why it took 10 years for process optimization with the vast majority of these studies using one specific material i.e. polycaprolactone (PCL).

Earlier studies achieved printing fidelity with MEW using an approach based on intuition, i.e., manually selecting values for the critical process parameters, performing post-printing fidelity measurements, assessing trends and patterns in data, and selecting process parameter settings for follow-up experimentation. Later studies focused on understanding the previously identified printing regimes with respect to the physics and the dynamics of the process [8–10]. A recent study systematically approached the calibration process by exploring the parameter space using a Design-of-Experiments approach in a simple Cartesian grid defined by the number of independent process parameters [11]. In this study, computer vision was employed to image the jet in the free-flow regime as a function of various process parameter conditions in a high throughput manner [11]. The generated dataset was then assessed offline to identify high fidelity printability regimes [11]. However, selecting an exploration strategy implies picking a resolution without knowing the model function. To address that, the resolution is often chosen high, aiming for an exhaustive search to avoid inaccuracies. With the high dimensionality of the parameter space, this brute force data collection method quickly fails to explore the space efficiently and becomes prone to bias.

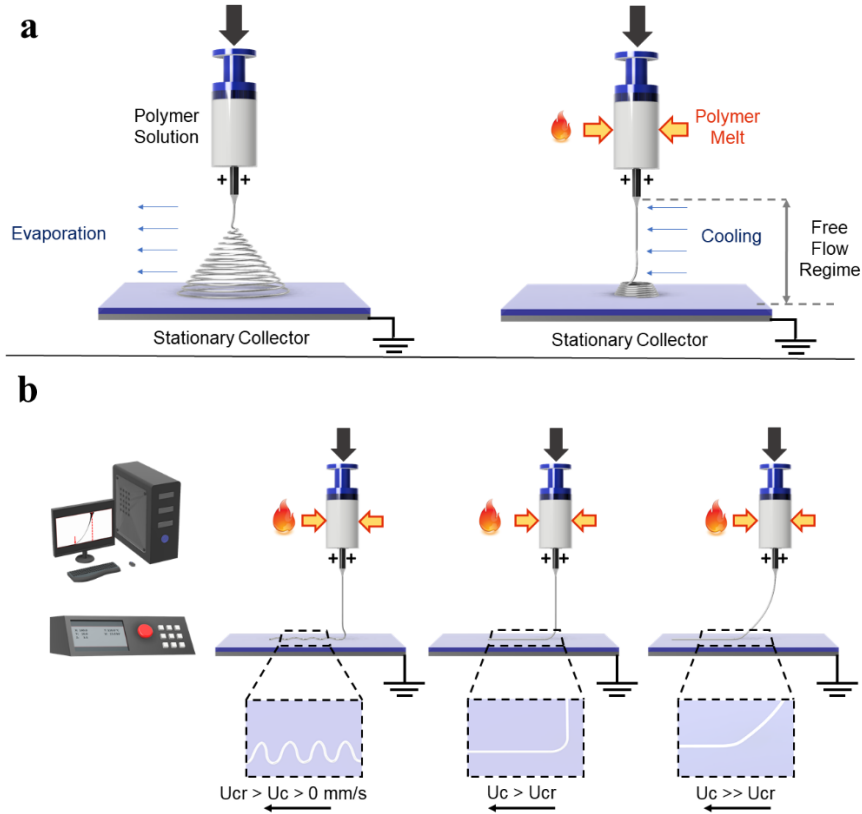

**Figure 1: Electrohydrodynamic Jet Printing Process.** Solution electrospinning (SES) vs. melt electrospinning (MES). The main differentiating feature between the two processes is the extent of the jet instabilities that arise from the electrostatic forces acting at the polymer jet-air interface. For MES, the chaotic jet regime is limited close to the grounded collector plate due to the high viscosity and dielectric properties of the pure polymer melt. b Direct melt electrowriting (MEW) and its operating principle.

The challenges mentioned before, combined with the demand for increasingly complex and reproducible products, warrant a new paradigm for E-jet printing machines. In this paradigm, rigid machines calibrated by trial-and-error practices are replaced by "intelligent" autonomous machines capable of adapting and learning process dynamics with minimum experimental cost. Artificial intelligence and Machine Learning (ML) are transforming many areas of experimental science in this direction. However, advances in manufacturing science are mainly driven by expensive physics-based simulations that cannot resolve all scales and, more recently, by data-hungry neural networks trained offline with in-process monitoring datasets for defect detection and process performance prediction on various AM platform technologies [12].

To address these challenges, we adopt an approach inspired by the operating principles behind autonomous materials experimentation platforms, also known as research robots [13–15] and from the field of physics-informed machine learning [16,17]. Research robots demonstrate closed-loop control through online learning from prior experiments, planning and execution of new experiments. Physics-informed machine learning lays the foundations for integrating data with domain knowledge in the form of mathematical models to allow efficient simulations of highly multi-physics phenomena. The underlying framework of research robots provides a systematic data-driven approach for the identification of the best follow-up experiments to optimize unknown functions. The functions are approximated by Gaussian Process Regression (GPR), which is a robust statistical, nonparametric technique both for function approximation and uncertainty quantification [18,19]. During the Bayesian optimization loop, an acquisition function balances the utilization of experiments that explore the unknown function with experiments that exploit prior knowledge by considering the quantified uncertainty after each function approximation step [20]. Efficiency with respect to the utilization of experimental resources could be further improved by augmenting the surrogate model with prior domain knowledge following a multi-fidelity modeling approach [21–23]. The success of this approach has been documented in the field of computational science by using simple and potentially inaccurate models that carry a low computational cost to achieve predictive accuracy on a small set of high-fidelity observations obtained from accurate models that carry a high computational cost.

Automated materials experimentation systems driven by Bayesian optimization active learning frameworks have demonstrated remarkable performance in autonomously searching the vast synthesis-process-structure-property landscape resulting to the accelerated discovery of advanced materials for a wide variety of applications [20,24–27] including AM[28]. However, their application not only towards the calibration of E-jet printing processes, but also in general AM technologies remains significantly underexploited. In one study concerning E-jet printing of substrates with micron-scale topographical features, the authors demonstrated a research robot, whose planner is informed by an in-line nano-surface metrology tool and actively learns to tune the extrusion rate until it achieves a predefined topographical feature [29]. In another study about direct ink writing of paste materials, the authors demonstrated an autonomous 3D printer, whose planner is informed by machine vision cameras and adaptively searches the space of four process parameters to print single struts with geometrical features that match user-defined specifications [30].

In this paper, we develop and demonstrate GPJet, an end-to-end physics-informed probabilistic machine learning framework that sets the basis for the next generation of self-calibrating E-jet printing machines. We construct a virtual MEW machine using a previously published video dataset acquired by a conventional camera that performs in situ jet monitoring under various process conditions, and we demonstrate that GPJet is capable of:

- high-fidelity jet feature extraction in real-time from video data using a parallelized computer vision algorithmic workflow that is systematically profiled under various implementations,
- low-fidelity jet feature extraction from "cheap" physics-based models describing the evolution of the jet across the free-flow regime and the deposition dynamics of a gravity-driven viscous thread onto a moving surface known as the "fluid-mechanical sewing machine" and
- learning the process dynamics with minimum experimental cost as described by the required number of high-fidelity data. This is supported by performance tests comparing offline and online calibration scenarios revealing that the online ML planner can effectively learn the jet evolution in the free-flow regime much more efficiently when it is informed by physics and based on that to adaptively tune the translational speed of the collector for minimum jet lag distance. In that case, the ML planner follows a decision-making strategy revealing the universality of the fluid mechanical sewing machine model in predicting the deposition dynamics of any printing process of viscous jets no matter what the nature of the jet driving force is.

This paper is organized as follows: First, we introduce GPJet, the physics-informed machine learning pipeline modules that we developed. Then, we describe the curated dataset we used to demonstrate the pipeline's utility and performance using MEW as a testbed. Third, we describe the results of our study, starting with the real-time jet feature extraction and how these are leveraged by the ML planner within GPJet for batch and online learning process dynamics using data and physics. Then, we discuss the advantages of GPJet as a platform whose utility can be generalized for any AM process and its current limitations. Lastly, we conclude with the current work in progress to realize fully autonomous closed-loop ML-driven E-jet printing manufacturing platforms.

## 2. GPJet: The Physics-Informed Machine Learning Pipeline

To demonstrate the ability of learning the dynamics of E-Jet printing processes in a data-driven fashion, we employ a pipeline-based approach that is depicted in **Figure 2**. The approach is composed of three modules, namely: the Machine Vision module, the Physics-based Modeling Module, and the Machine Learning Module. In GPJet, features that are representative of the printing process dynamics, are extracted by the Machine Vision module and the Physics-based modeling module. In the context of this paper, high-fidelity observations are referred to the jet features extracted experimentally, and low-fidelity observations are referred to the same jet features as predicted by a low-cost numerical model that is a good approximation of the reality.

As a first step, jet features are engineered and extracted in real-time using an algorithmic computer vision workflow taking as an input time-series video data (see Methods for details). The Machine Vision module allows us to probe and measure the jet dynamics, a capability hereafter denoted as jet metrology. The jet metrology serves as a feature extraction step of high-fidelity observations corresponding to the jet radius profile ($R_j$ [mm]) and the jet lag distance ($L_j$ [mm]), which are then fed into the Machine Learning module that can perform various Bayesian-based batch and online learning tasks (see Methods for details). The Machine Learning module can be further informed by low-fidelity observations, a capability hereafter denoted as Multi-fidelity modeling. The low-fidelity observations are obtained by the Physics-based modeling module and correspond to the same engineered features that are extracted experimentally by the Machine Vision module ($R_j$ [mm] and ($L_j$ [mm]).

Collectively, the GPJet pipeline offers a range of unique capabilities ranging from real-time feature extraction using computer vision to physics-informed machine learning capabilities that aim to minimize experimental cost without sacrificing accuracy and robustness.

## 3. Dataset

To demonstrate the utility and performance of the GPJet pipeline, we curated a dataset that emulates a virtual E-jet printing machine with a dynamic range of 12 user-controlled machine settings. The dataset is depicted in **Table 1** and is created based on previously published time-series video data [10]. Specifically, the raw data is acquired by a conventional camera with 50 fps and a field of view spanning the area between the needle tip and the grounded collector of a melt electro-writing (MEW) system. A detailed explanation of the raw data, the preprocessing procedure derive the final curated dataset can be found in **Supplementary Note**. MEW constitutes an ideal testbed for demonstrating the capabilities and the flexibility of our GPJet framework. The highly dynamic nature of the process and the multiple user-controlled independent process parameters, pose several challenges that we demonstrate both in an offline and an online self-calibrating machine scenario.

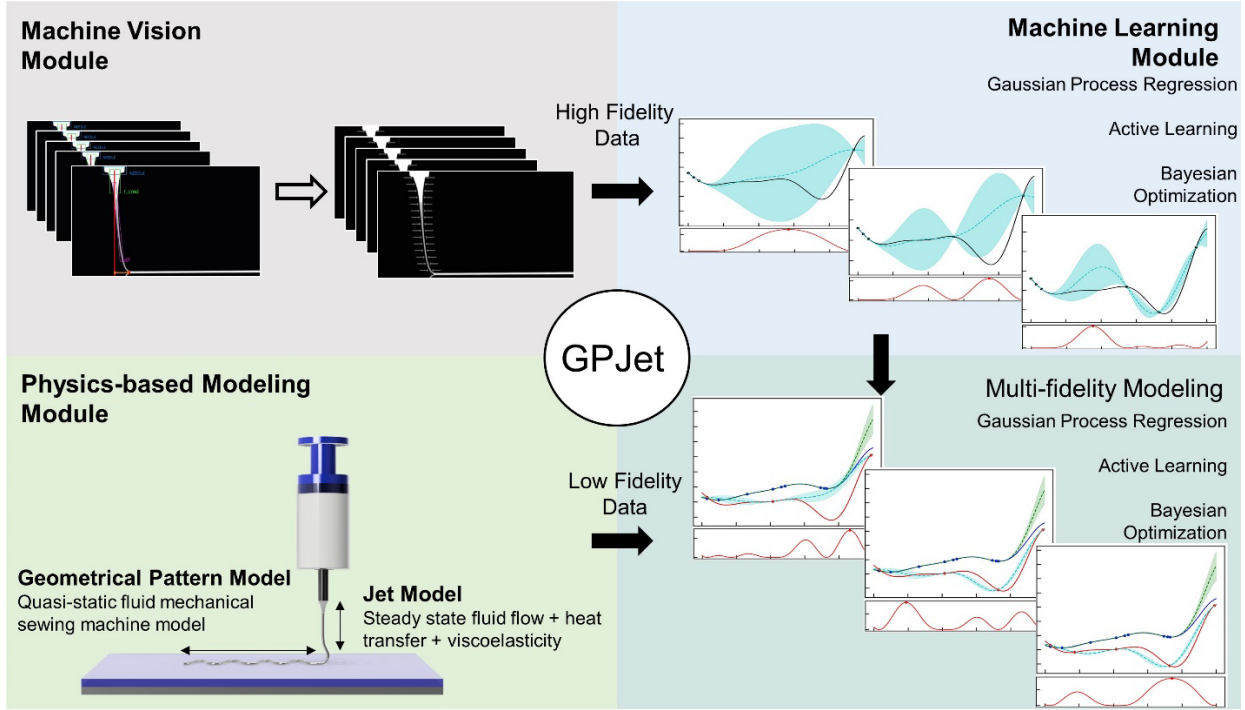

**Figure 2: The GPJet Pipeline Framework.** A Physics-informed Bayesian Machine Learning framework comprised by three different modules: a) the Machine Vision module, which takes as an input timeseries video focusing on the polymer jet in the free flow regime and performs extraction of high-fidelity jet features in real-time based on an automated image processing workflow b) the Physics-based Modeling Module, which and c) the Machine Learning module, which takes as an input high fidelity experimental data from the Machine Vision module and low fidelity modeling data from the and performs a series of data-driven tasks to learn the jet dynamics.

**Table 1**: Overview of curated dataset that emulates an E-jet printing machine.

| Machine Setting | Air pressure $p$ [$bar$] | Tip to Collector Distance $Z$ [$mm$] | Collector Speed $U_c$ [$mm/s$] | Number of frames | Duration [$sec$] |
|---|---|---|---|---|---|
| 1 | 1.2 | 3.5 | 191.2 | 1341 | 26.82 |
| 2 | 1.2 | 3.5 | 212.5 | 1672 | 33.44 |
| 3 | 1.2 | 3.5 | 255 | 1437 | 28.74 |
| 4 | 1.2 | 3.5 | 340 | 1343 | 26.86 |
| 5 | 1.2 | 3.5 | 510 | 648 | 12.96 |
| 6 | 1.2 | 3.5 | 850 | 613 | 12.26 |
| 7 | 1.2 | 3.5 | 1530 | 457 | 9.14 |
| 8 | 1.2 | 3.5 | 2890 | 401 | 8.02 |
| 9 | 2.4 | 4.5 | 292.5 | 1108 | 22.16 |
| 10 | 2.4 | 4.5 | 520 | 802 | 16.04 |
| 11 | 2.4 | 4.5 | 1300 | 812 | 16.24 |
| 12 | 2.4 | 4.5 | 4420 | 284 | 5.68 |

See Supplementary Note for data source and pre-processing.

## 4. Results

### 4.1 Learning Jet Dynamics from Videos

As a first goal we set out to tackle the challenge of real-time process monitoring and jet metrology. To demonstrate the highly dynamic nature of the process, we plot overlaid video frames showing the jet hitting a stationary collector (**Figure 3a**). We chose to plot frames with a time step equal to 0.2 sec since the electrostatic nature of the process and the viscoelasticity of the molten jet cause instabilities of a significantly smaller time scale (~0.02 sec) and result in jet topologies that are indistinguishable with a naked eye. This number provided a starting point for setting a goal related to the computational efficiency of the machine vision module for real-time performance. Since the camera acquisition time was equal to 0.02 sec (50 fps), we proceeded with the goal to maintain computational processing time equal or smaller than that.

To accomplish this, we started by dividing the computer vision workflow in specific algorithmic tasks and implemented a sequential code version. We continued by systematically profiling the code, identifying the computationally expensive tasks, and then gradually parallelizing the code to reduce computational processing time. This approach led to three different code implementations of the machine vision module: a) the sequential, b) the concurrent and the c) parallel, with the last one achieving real-time performance. The results of the profiling experiments are shown in **Figure 3b**, where all the tasks are plotted along with their respective processing time for the three different code implementations.

Specifically, the machine vision tasks per frame are the following:
Task 1: Read new video frame.
Task 2: Process the frame to reverse background color.
Task 3: Edge-based feature extraction and data storage.

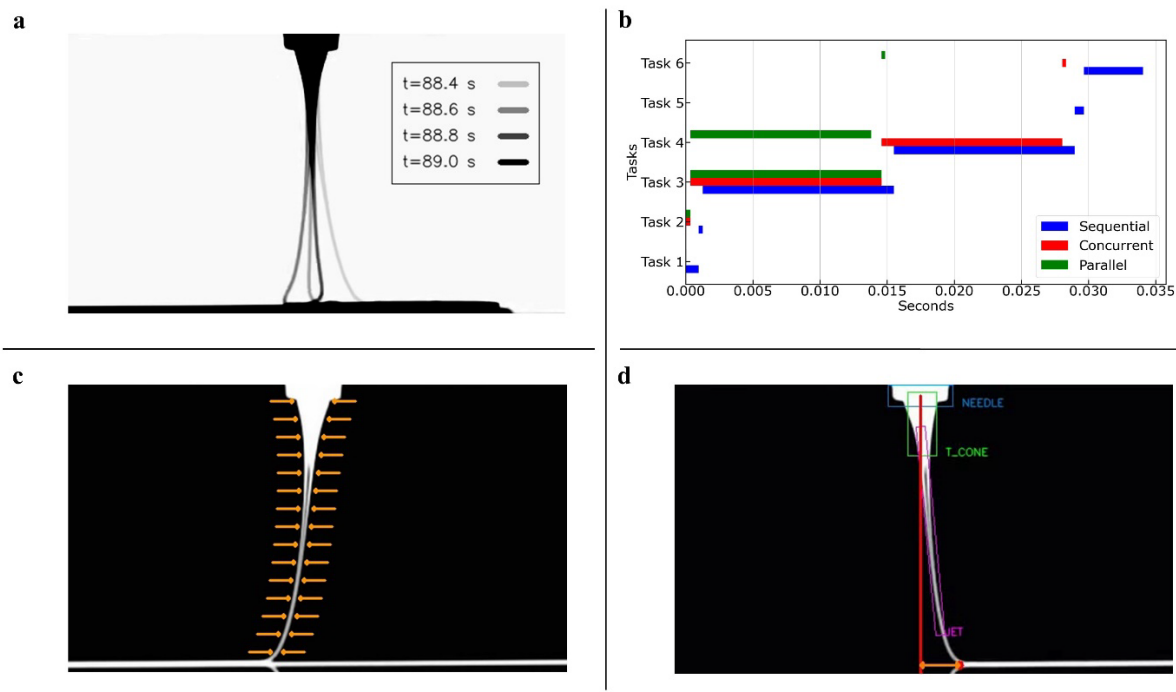

**Figure 3: Machine Vision Module. a)** Process dynamics and its scale. **b)** Profiling experiments for different code implementations. **c)** Edge-based feature extraction methodology (Task 3 in Figure 3b). **d)** Object-based feature extraction methodology (Task 4 in Figure 3b).

Task 4: Object-based feature extraction and data storage.
Task 5: Show processed video output.
Task 6: Save video output.

Profiling the sequential code version reveals that an average time of 0.033 sec. is needed to perform the whole machine vision workflow per frame with the most expensive task being the one that performs edge-based feature extraction across the jet length (**Figure 3c**). To alleviate this source of computational cost, we employed a multi-threading strategy for the concurrent code version that led to a modest improvement of 0.005 sec.

Multi-threading is implementing software to perform two or more tasks in a concurrent manner within the same application. Multi-threading employs multiple threads to perform each task with no limitation in the number of threads that can be used [10]. We learned that multithreading on one hand can reduce processing time of I/O bound tasks almost to zero, but on the other hand does not improve processing time of CPU bound tasks, such as Task 3 and Task 4, which are the most expensive.

To further reduce processing time, we augmented the concurrent version with a multi-processing strategy that led to the parallel code version. Multi-processing systems have multiple processors running at the same time. Therefore, different tasks of an application can be run in different processors in a parallel manner. This capability considerably accelerates program performance. The limitation of this strategy is related to the fact that the number of processes that can be employed must be less or equal to the number of processors (CPU cores) of the device [10]. Finally, by employing multi-threading for I/O bound tasks (Task 1, Task 5 and Task 6) and multi-processing for CPU bound tasks (Task 3, Task 4), we were able to achieve real-time process monitoring and jet metrology with processing time up to 0.014 sec.

Instrumented with the capability to perform jet feature extraction in real-time, we then focused on quantifying process dynamics relevant features. With the edge-based feature extraction algorithm, which is described in detail in sub-section 6.2 under the Methods section, we were able to measure the jet diameter profile, the area of the whole jet, the angle between the vertical line that connects the nozzle tip with the collector, and different points across the length of the jet profile and finally the translational jet speed at different points across the length of the jet profile. The high content spatio-temporal results are plotted in **Figure 1** of the **Supplementary Information** demonstrating the breadth of information of the machine vision module and the fact that the jet point right above the collector undergoes a highly fluctuating behavior that will directly affect the printing quality.

We present the jet metrology results for two distinct phases during the printing process in **Figure 4ai-ii** and **Figure 4bi-ii** focusing on the jet point right above the collector, hereafter denoted as point of interest. With the object-based feature extraction algorithm which is described in detail in sub-section 5.2 under the Methods section, we were able to detect key objects in the field of view such the needle tip, the Taylor cone, which is defined as the jet area between the needle tip outlet and the jet point 2Ro away from the needle tip, the remaining jet and the collector. In this way, we were able to measure the Lag distance, defined as the distance between the point of interest and the projection of the middle point of the nozzle tip outlet to the collector. All detected objects are denoted graphically in **Figure 3d**, which shows the video output after Task 4 during the computer vision workflow.

As a next step, we asked how we could leverage the extracted features to learn the dynamics of the process in the most efficient data-driven way, with respect to both experimental and computational cost. To address this question, we developed several Bayesian learning techniques, hereafter denoted as the Machine Learning module of the GPJet framework. The Machine Learning module takes as input the extracted high-fidelity data and initially uses Gaussian Processes (GPs) to approximate the function describing the relationship between a) the jet radius profile and the nozzle tip to collector distance and b) the Lag distance and the ratio of the collector speed over the jet speed at the point of interest.

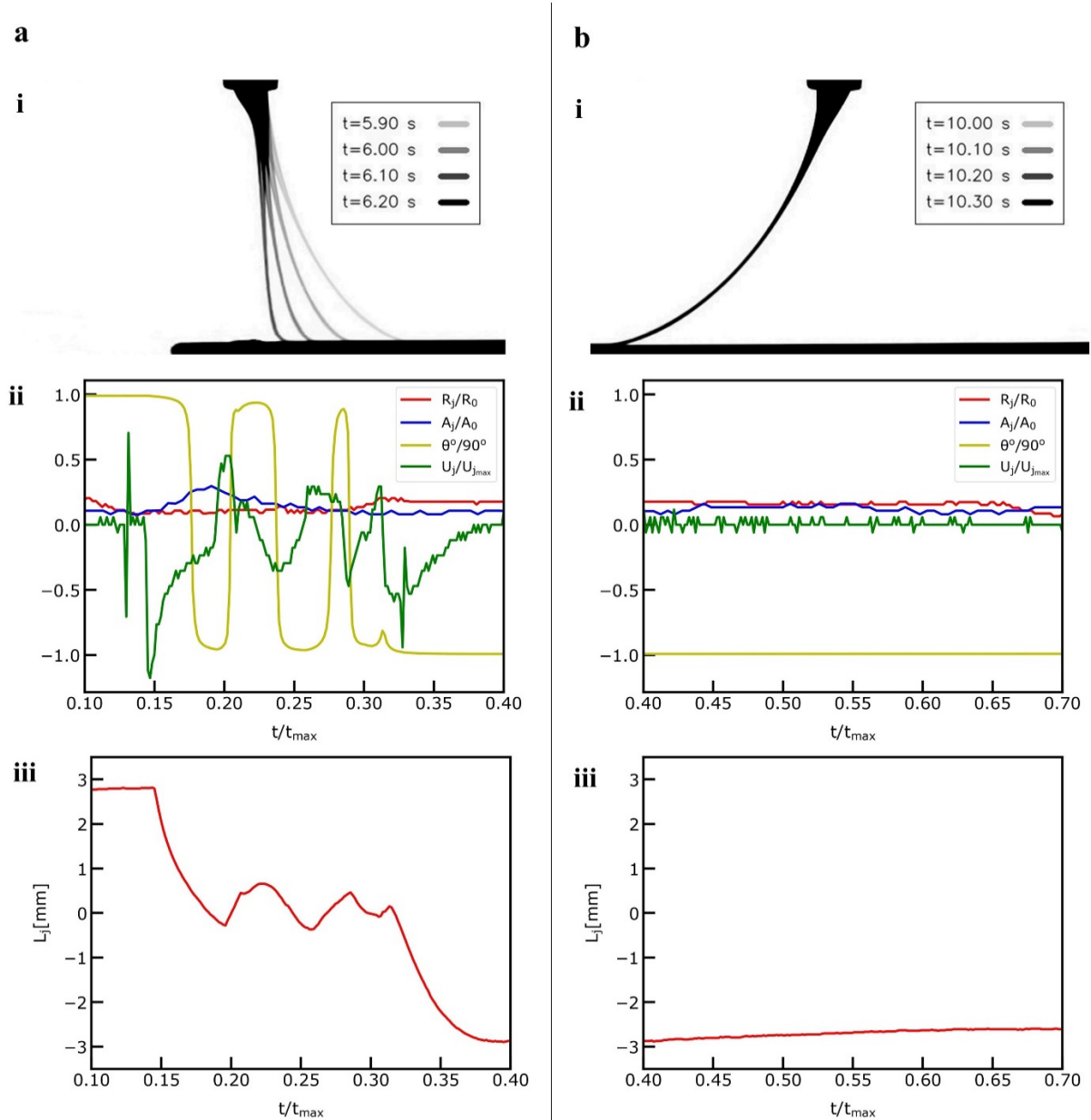

**Figure 4: Jet Metrology with the Machine Vision Module. a)** The extracted features during the deceleration-acceleration phase of the printing process. **i)** Overlayed video frames demonstrating the dynamics during the deceleration-acceleration phase and normalized jet length point of interest ($Z/R_o = 17.5$) denoted with red color. **ii)** Normalized jet radius ($R_j/R_o$), Normalized jet area ($A_j/A_o$), Normalized jet angles ($\theta/90^o$) and Normalized jet velocity ($U_j/U_{jmax}$) at the denoted point of interest plotted against the normalized time ($t/t_{max}$) during the deceleration-acceleration phase. **iii)** Jet lag distance ($L_j$) plotted against the normalized time ($t/t_{max}$) during the deceleration-acceleration phase. **b)** The extracted features during the steady speed phase pf the printing process. **i)** Overlayed video frames demonstrating the dynamics during the steady speed phase. **ii)** Normalized jet radius ($R_j/R_o$), Normalized jet area ($A_j/A_o$), Normalized jet angles ($\theta/90^o$) and Normalized jet velocity ($U_j/U_{jmax}$) at the denoted point of interest plotted against the normalized time ($t/t_{max}$) during the steady speed phase. **iii)** Jet lag distance ($L_j$) plotted against the normalized time ($t/t_{max}$) during the steady speed phase.

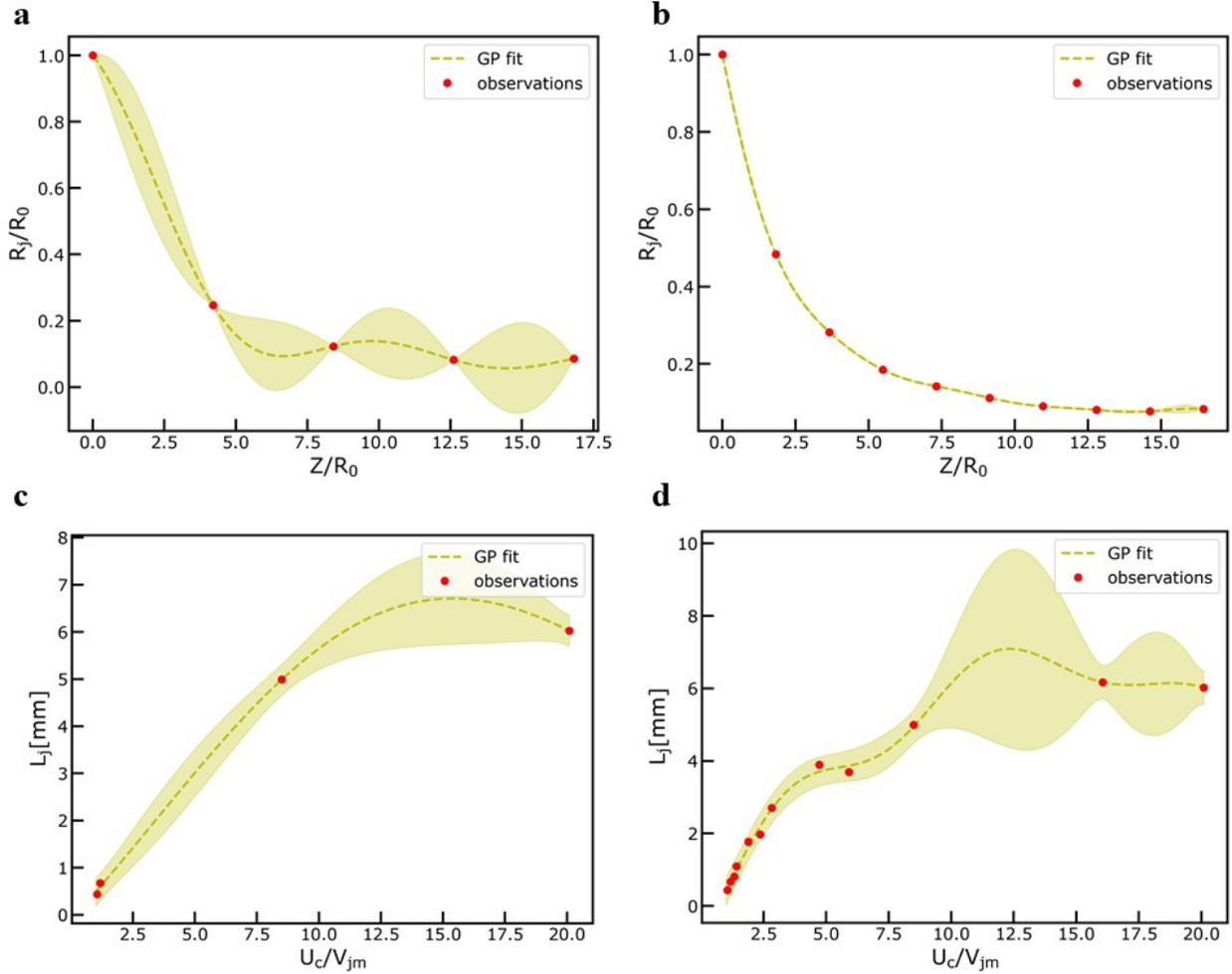

**Figure 5**: **Results of Gaussian Process Modeling Regression Tasks**. **a**) fitting normalized ($R_j/R_o$) jet radius observation data (n=5) obtained from the computer vision metrology module of the GPJet framework at specific z axis coordinates along the normalized jet length ($Z/R_o$). **b**) fitting normalized jet radius using a higher number of observation data (n=10) compared to the previous case (a). **c**) fitting lag distance ($L_j$) observation data (n=3) obtained from the computer vision metrology module of the GPJet framework for specific speed ratios ($U_c/V_{jm}$). **d**) fitting lag distance using all available observation data (n=12). For non-normalized quantities units are in SI. Filled contours represent uncertainty bounds (95% confidence intervals (CIs)).

Gaussian process regression (GPR) is a robust statistical, non- parametric technique for function approximation with kernel machines. GPR provides the important advantages of uncertainty quantification, the ability to perform well with small datasets and the capability to easily include domain-aware physics-based models in the deployed kernels.

To learn how the jet radius profile evolves over the tip to collector distance, we chose radial basis functions (RBF) as the kernel approximator and performed GPR. We trained the model under two different scenarios with n=5 observations and n=10 observations chosen at equally spaced points along the jet length for the $1^{st}$ and $2^{nd}$ scenario, respectively. It is important to mention that the machine vision module provides n = 93 observations along the jet length. The results are shown in **Figure 5a** and **Figure 5b** for the two different training scenarios. GPs can approximate the jet radius profile evolution with just n = 10 observations showcasing the efficiency of our data-driven approach with respect to computational cost.

To learn the function describing the relationship between the Lag distance and the ratio of the collector speed over the jet speed at the point of interest, we employ the same modeling strategy as before. Similarly, we set up two different training scenarios with n=4 observations and n=12 observations, respectively. Please note here that the number of high-fidelity observations at our disposal is constrained by our previously published experimental dataset (see sub-section 5.1 under the Methods section), where videos were acquired only at 12 different speed ratio settings. The results are shown in **Figure 5c** and **Figure 5d** for the two different training scenarios. While in the $1^{st}$ training scenario, GPR provides a smooth function approximation, the prediction's error from the experimental ground truth quantified by the Root Mean Square Error (RMSE), is significantly higher compared to the $2^{nd}$ training scenario (**see Figure 3-b-d in Supplementary Information**). As a result, the function describing the relationship under question, is hard to approximate due to the high variability of the Lag distance caused by the jet instabilities close to the collector.

Collectively, our machine vision module informing the GPR capabilities of the machine learning module with high-fidelity observations demonstrates that we can learn the dynamics of the process. Specifically, GPJet demonstrates excellent performance with respect to the prediction of jet radius profile evolution for a small amount of high-fidelity observations n = 10. Furthermore, GPJet demonstrates very good performance for the available number of high-fidelity observations with respect to the Lag distance behavior at different collector speed settings.

**4.2 Learning Jet Dynamics from Videos & Physics**

As a next step, we focused on exploring how we could further reduce the number of high-fidelity observations without losing the predictive capability of GPR with respect to the jet radius profile evolution. To accomplish that, we augmented the high-fidelity observations obtained by the machine vision module with low-fidelity observations obtained in a principled manner by a multi-physics model. The multi-physics model captures the electro-hydrodynamics, the heat transfer and viscoelastic constitutive material behavior of the molten jet in 1D across the needle tip to collector distance. The mathematical formulation and numeric implementation of the model are described in detail in sub-section 5.3 under the Methods sections.

We set up our data-driven scheme with two fidelities corresponding to two different kernel machines integrated in one multi-fidelity kernel, in which the correlation between the two kernels is encoded as a linear relationship. In other words, we constrain the prior knowledge during GPR with physics-relevant knowledge, resulting to a physics-informed posterior prediction that requires much less high-fidelity observations.

We trained the multi-fidelity model under two different scenarios with n=6 high-fidelity observations and n=7 high-fidelity observations, respectively. For both scenarios the number of low-fidelity observations was kept to a number equal to 32 and equally spaced points across the jet length. For the $1^{st}$ scenario n=6 equally spaced points were chosen across the jet length depicted in the jet schematic of the **Figure 6a** (upper-left). The results are shown in **Figure 6a-i** and **Figure 6a-ii.** In **Figure 6a-i,** we plot the multi-fidelity GPR predictions for the low and high-fidelity observations respectively. In **Figure 6a-ii**, we plot the predictions of the multi-fidelity GPR in high-fidelity observations together with the predictions of a simple GP in high-fidelity observations. Both plots demonstrate that we can learn the jet radius profile much better using two different fidelities compared to using only one fidelity for the same number of high-fidelity observations. Our results, point out that we lose predictive accuracy for the Taylor cone area (below the needle tip outlet). This phenomenon was expected due to that the fact that similar behavior was observed when the multi-physics model was tested and informed the strategy of the 2$^{nd}$ scenario, where we chose 7 high-fidelity observations with the additional point being in the Taylor cone area. The results are shown in **Figure 6b-i** and **Figure 6b-ii** demonstrating that we have managed to further reduce the required number of high-fidelity observations that need to be extracted by the machine vision module without compromising the predictive accuracy.

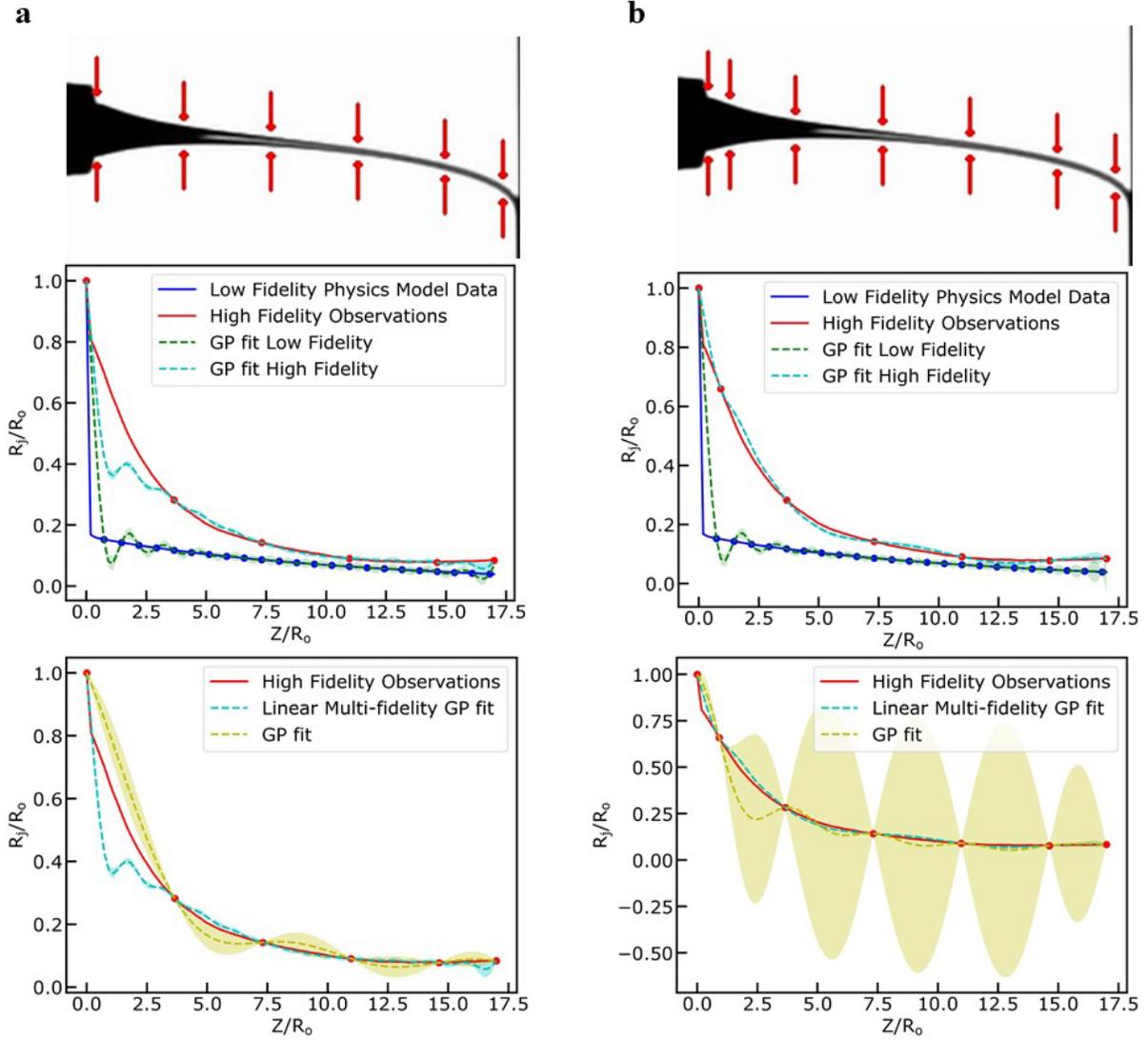

**Figure 6**: **Results of Multi-fidelity Modeling Regression Tasks. a)** fitting normalized high fidelity observation data (n=6, red color) of jet radius ($R_j/R_o$) and low fidelity model data obtained from the computer vision metrology module of the GPJet framework and from the multi-physics model, respectively, at specific z axis coordinates along the normalized jet length ($Z/R_o$) and comparing the results with a simple GP fit using the same number of high fidelity observation data. **b)** fitting a higher number of normalized high fidelity observation data (n=7, red color) of jet radius ($R_j/R_o$) and low fidelity model data obtained from the computer vision metrology module of the GPJet framework and from the multi-physics model, respectively, at specific z axis coordinates along the normalized jet length ($Z/R_o$) and comparing the results with a simple GP fit using the same number of high fidelity observation data.

### 4.3 Active Learning of Jet Dynamics

Up to now, we demonstrated that GPJet, is a robust tool for passive learning of jet dynamics. By "passive", we mean that given a high-fidelity dataset provided by the Machine Vision module and augmented by low-fidelity data provided by the Physics-based module, the GPR capabilities of the Machine Learning module can model the function that mathematically represents the relation between the jet radius and the needle tip to collector distance. In addition to that, we employed the same strategy without low fidelity data, to model the function describing the highly dynamic relationship between the Lag distance and the ratio of the collector speed and the jet velocity at the point of interest.

In this section, we asked the questions of whether we could actively choose data points across jet length for which to observe the outputs to accurately model the underlying function describing the jet dynamics with respect to the extracted jet features. To accomplish that, we deploy a virtual MEW machine, whose dynamic range is defined by the available dataset, and we run simulation experiments to demonstrate if we can learn the underlying functions in an active manner as quickly and accurately as possible.

To accomplish that, we set up an exploration scenario, a set-up closely related to optimal experimental design scenarios as it equates to adaptively selecting the input spatial points across the jet length based on what is already known about the function describing the jet radius profile and where knowledge can be improved. We run active learning in both the multi-fidelity GP and simple GP for the jet radius profile evolution. The results are shown in **Figure 7**. To systematically, compare the performance of the two different models, we chose the same initial training points (**Figure7a-i** and **Figure7b-i**) and the same number of iterations during each training phase. For each iteration (**Figure 7a(i-vi)** and **Figure 7b(i-vi)**), we graphically show, on the processed video frame the adaptively selected point across the jet length and below that the modeling results. The adaptive selection is based on a purely exploratory acquisition function that steers the point selection towards the area of least knowledge quantified by the uncertainty output of the modeling step. The results demonstrate that we can learn actively and in a purely exploratory scenario accurately and fast the underlying function. Each iteration phase for the multi-fidelity (MFD) GPs and simple GPs lasts around ~ 0.5 seconds leading to a total learning time equal to 3 sec. Lastly, we extract performance metrics to compare the active learning between the multi-fidelity and simple GP model (see **Figure 2-a-c** in **Supplementary Information**). The results demonstrate that active learning on the MFD model is significantly faster (**Figure 2-a-c**) with more confident predictions since the model's prior assumptions are constrained by domain-aware data.

Then, we employ the same strategy to actively learn the function describing the relation between the Lag distance and the speed ratio (put symbol) in an exploration scenario. The results are shown in **Figure 8**. The virtual MEW machine performs remarkably well in the prescribed experimental simulation. It starts by randomly selecting one speed ratio equal to 5 (see **Figure 8-a**) and after 4 additional iterations (see **Figure 8-a-b-c-d**), the underlying function is quite effectively approximated. Performance metrics (see **Figure 3-b-d** in **Supplementary Information**) demonstrate that the underlying function can be learned fast in an active manner

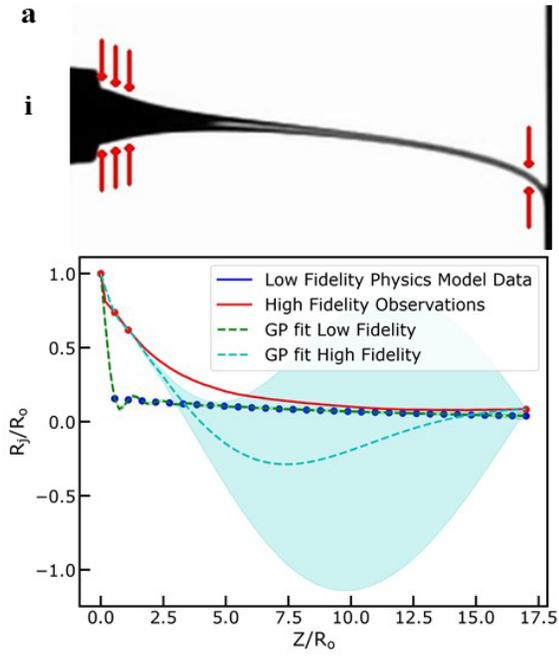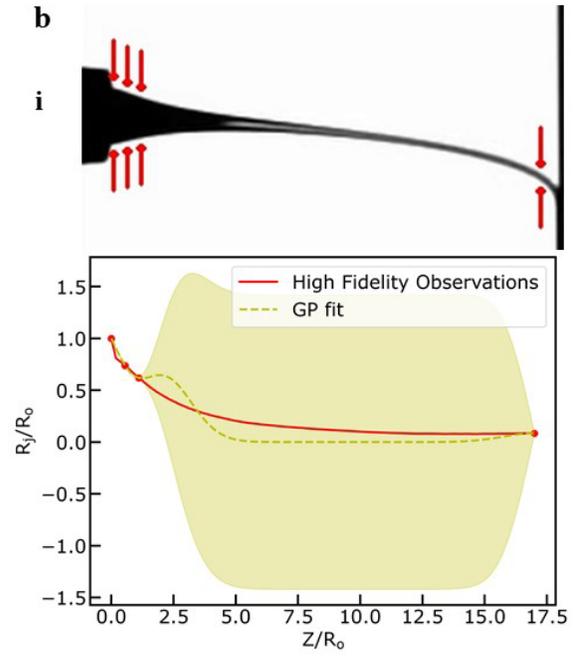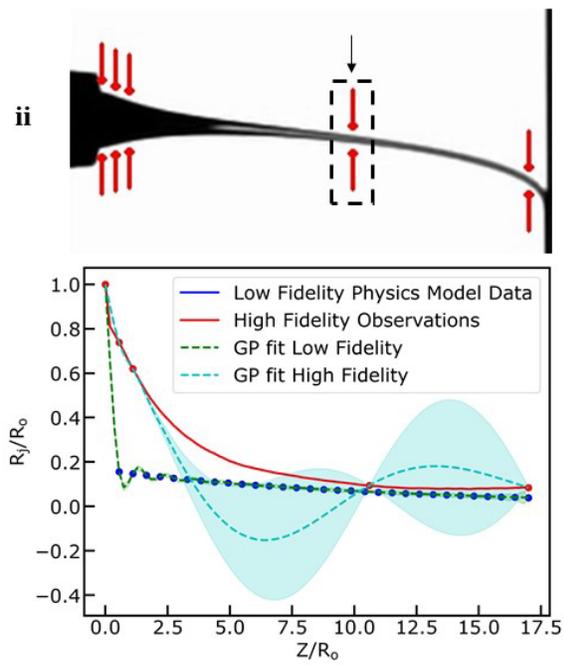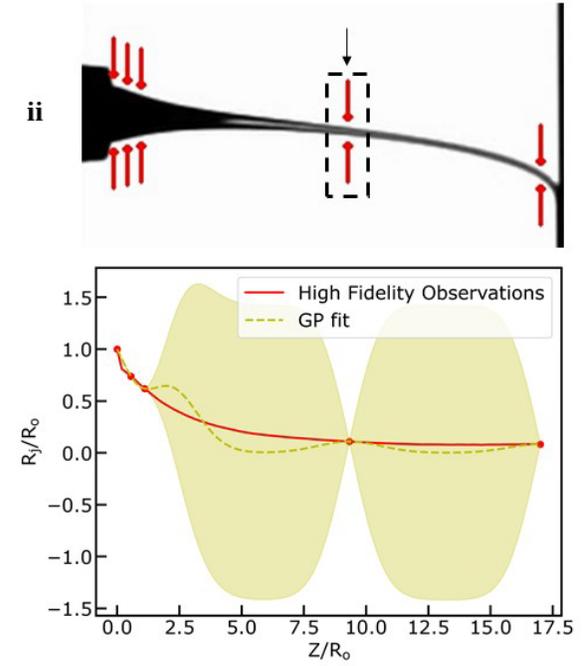

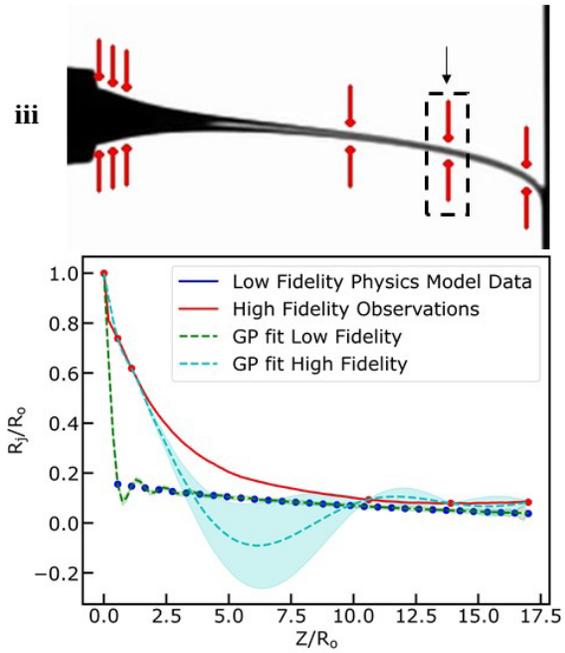
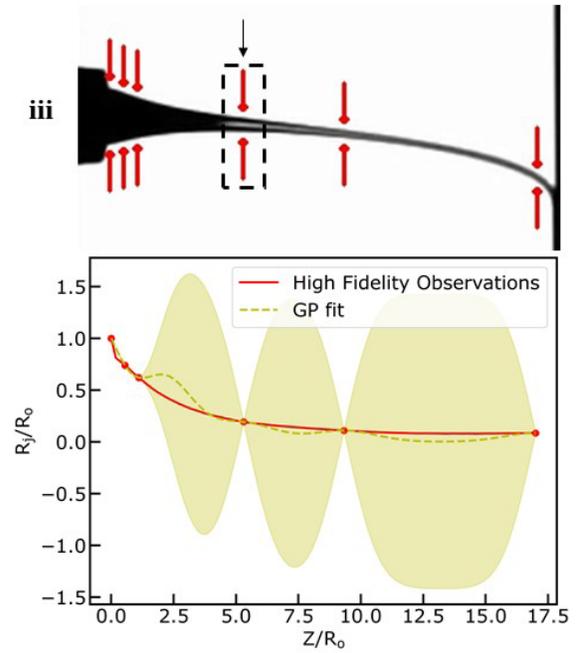
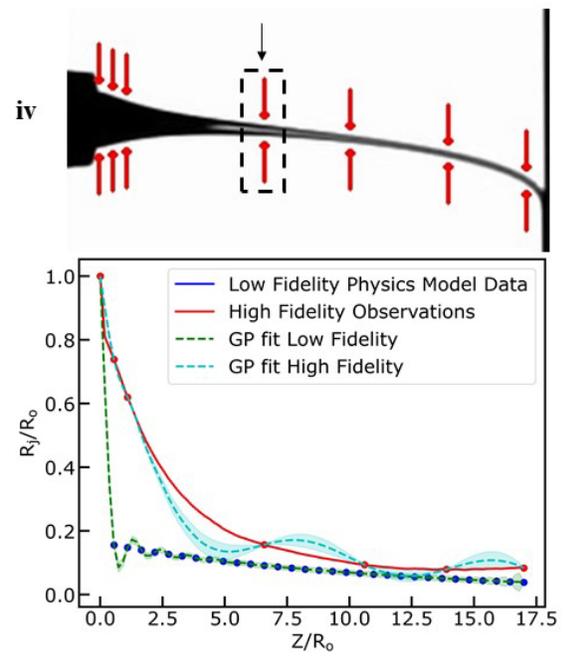
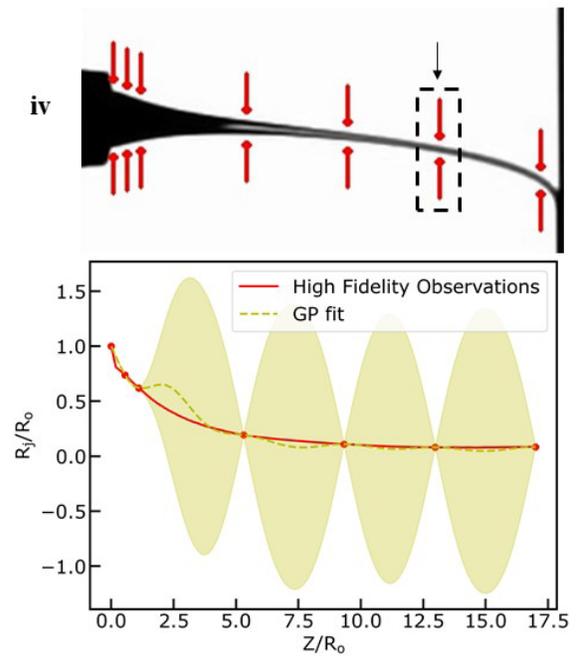

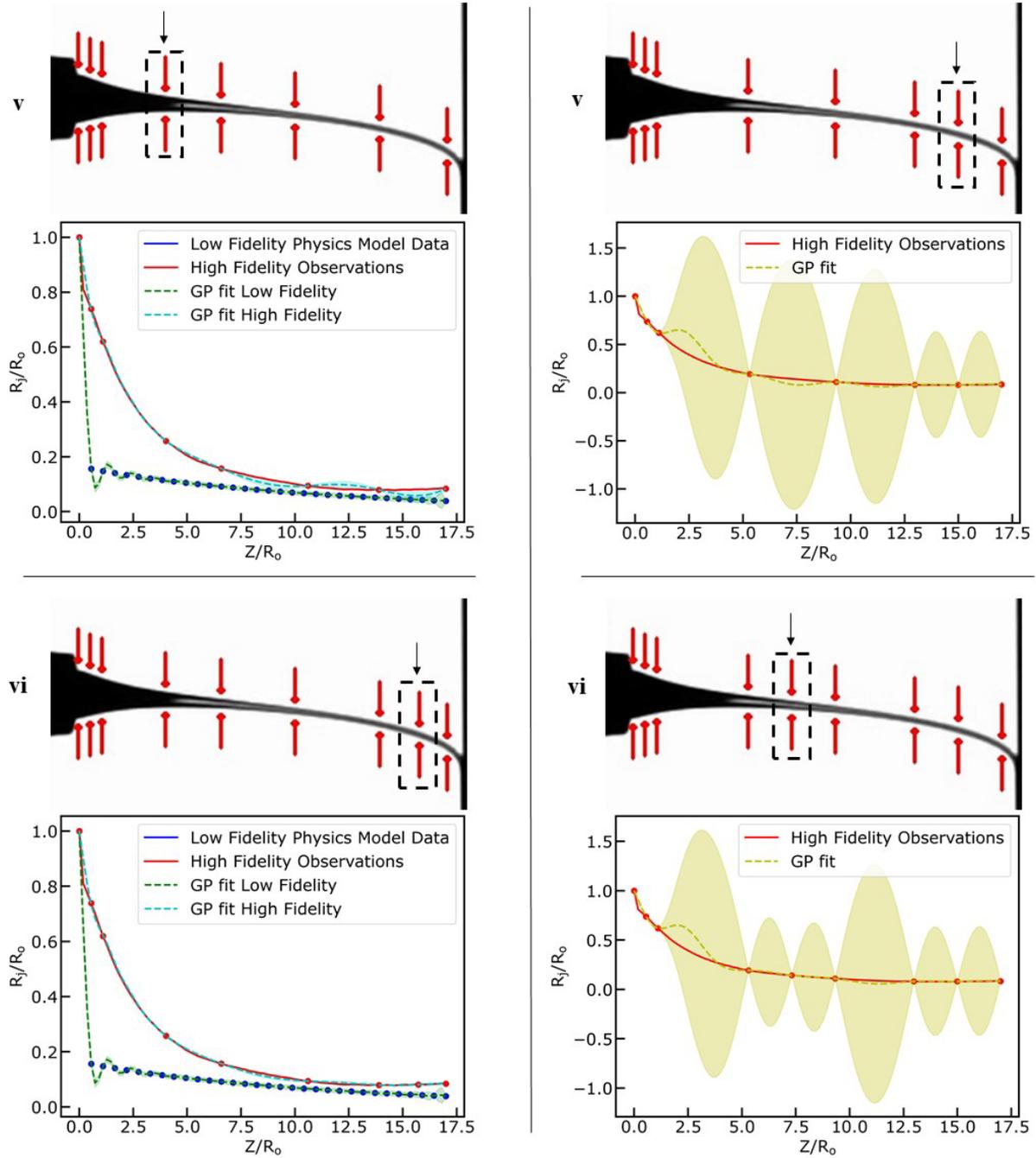

**Figure 7: Results of Active Learning process on Multifidelity Modeling Task. a)** exploring the design space using Active Learning to fit a Multifidelity Gaussian Process to normalized high fidelity observation data (red color) of jet radius ($R_j/R_o$) and low fidelity model data obtained from the computer vision metrology module of the GPJet framework and from the multi-physics model, respectively, at specific z axis coordinates along the normalized jet length ($Z/R_o$) ( **i – vi** denote the iterations of the active learning algorithm until it meets its termination criteria). **B)** exploring the design space using Active Learning to fit a Gaussian Process to normalized high fidelity observation data (red color) of jet radius ($R_j/R_o$) obtained from the computer vision metrology module of the GPJet framework at specific z axis coordinates along the normalized jet length ($Z/R_o$) ( **i – vi** denote the iterations of the active learning algorithm until it meets its termination criteria).

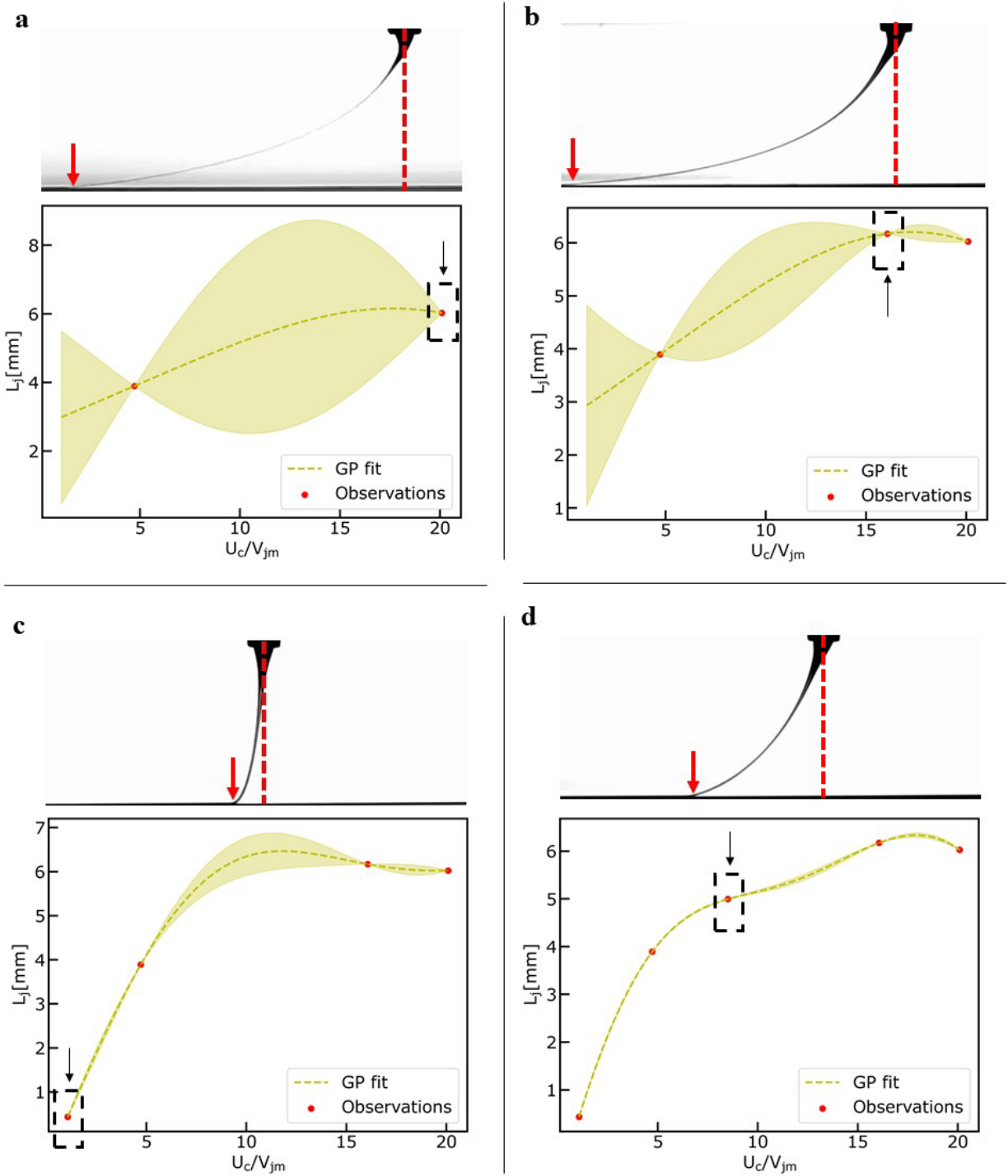

**Figure 8: Results of Exploring the Design Space Task.** Exploring the design space using active learning to fit a Gaussian Process Model to lag distance ($L_j$) observation data obtained from the computer vision metrology module of the GPJet framework for specific speed ratios ($U_c/V_{jm}$). **a-d)** Iterations of the active learning algorithm until it meets termination criteria.

and provide predictions with higher confidence compared to the passive learning approach and

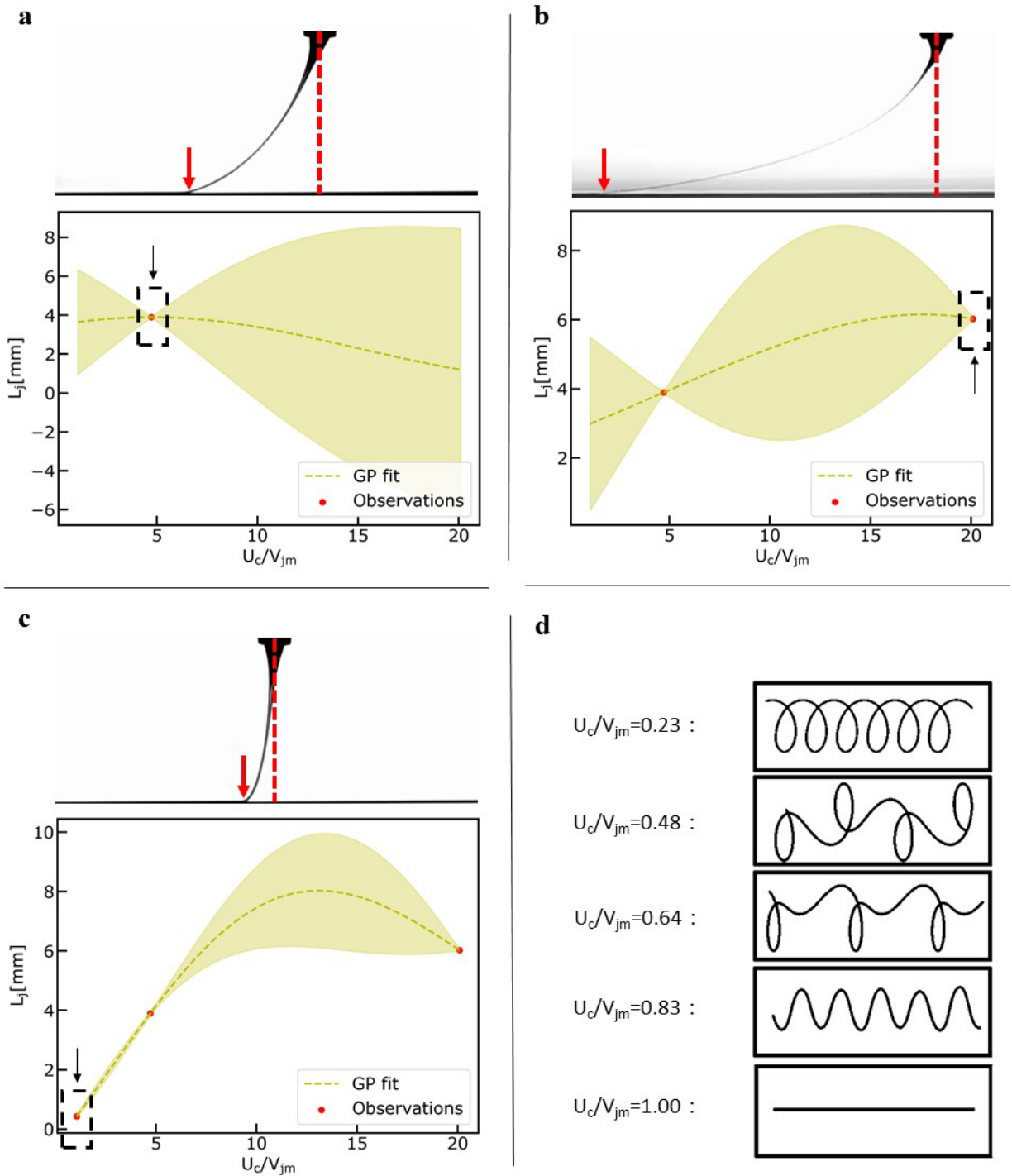

**Figure 9: Results of Bayesian Optimization Task.** Performing Bayesian Optimization to find the minimum lag-distance ($L_j$) by fitting a Gaussian Process Model to lag distance ($L_j$) observation data obtained from the computer vision metrology module of the GPJet framework for specific speed ratios ($U_c/V_{jm}$). **a-c)** Iterations of the Bayesian optimization algorithm until it meets termination criteria. **d)** For speed ratios less than one ($U_c/V_{jm} < 1$) the process is unstable, no straight line is formed, instead the translated coiling, alternating loops, W patterns and meanders patterns are formed, therefore no lag distance ($L_j$) observation data can be obtained from the computer vision metrology module of the GPJet framework.

Finally, we set out to address the following question. Can the virtual MEW machine find the speed ratio corresponding to the minimum Lag distance in an autonomous way? Autonomy in this paper, refers to the machine's ability to self-drive measurements of an experiment. Some initial parameters, such as the parameters to explore and their corresponding ranges constrained by the dataset, is defined by the user a priori. Instead of us learning the relation between the Lag distance and the speed ratio and afterwards calibrating the machine hyperparameters, we aim to demonstrate a self-calibrating scenario. To achieve that we employ an exploitation-exploration stategy in the spirit of Bayesian Optimization (BO)$^{Lookman\,(2019)}$. It is called exploration–exploitation as scenarios where the output of the underlying function must be optimized require us to both sample uncertain areas to acquire more knowledge about the function (exploration) as well as sampling input points that are likely to produce extremum outputs given the current knowledge of the function (exploitation). The virtual MEW machine performs remarkably well in the prescribed experimental simulation. It starts again by randomly selecting a speed ratio equal to (see **Figure 9-a**) and after 2 additional iterations (see **Figure 9-a-c**) the speed ratio corresponding to the minimum Lag distance has been reached. This speed ratio is close to 1, as expected from the mechanical sewing machine model, which is described in detail in sub-section 6.4 under the Methods section. BO validates the initial hypothesis formed by universality about the mechanical sewing machine model.

## 5. Conclusions

In this work, we demonstrate GPJet, an end-to-end physics-informed probabilistic machine learning framework that sets the basis for the next generation of self-calibrating E-jet printing machines. We construct a virtual MEW machine using a previously published video dataset acquired by a conventional camera that performs in situ jet monitoring under various process conditions, and we demonstrate that GPJet is capable of:

- high-fidelity jet feature extraction in real-time from video data using a parallelized computer vision algorithmic workflow that is systematically profiled under various implementations,
- low-fidelity jet feature extraction from fast physics-based models describing the evolution of the jet across the free-flow regime and the deposition dynamics of a gravity-driven viscous thread onto a moving surface known as the "fluid-mechanical sewing machine"

and

- learning the process dynamics with minimum experimental cost as described by the required number of high-fidelity data. Two case studies were performed, one regarding the jet diameter profile and the other regarding the lag distance. We have proven, through Gaussian Process Regression, that for an offline learning strategy, the number of data and their respective position in the design space are crucial for the quality and the confidence of the predictions in both cases. Also, in the case of the jet radius profile, coupling high-fidelity data, provided from the machine vision module, with low-fidelity data, provided from the multi-physics numerical model, through multi-fidelity Gaussian Process Regression, can provide better and more confident predictions, while using less high-fidelity observations. Such results lead to computational cost reduction, since jet diameter needs to be evaluated in less points across the nozzle to bed distance, and hence to even faster video processing times. Then an online learning strategy was utilized. Bayesian Optimization algorithms were employed to actively learn the jet diameter profile, using the minimum experimental cost, with and without multi-fidelity modeling. Once more ML planner can effectively learn the jet evolution in the free-flow regime much more efficiently when it is informed by physics. Same strategy is employed.

## 6. Methods

### 6.1 Machine Vision Module

**Jet Metrology.** For the implementation of the Jet Metrology algorithm, Python 3.8 was used, along with the python bindings of the OpenCV library, which enables us to read and process video data. The Jet Metrology algorithm consists of two sub-algorithms. The first is the Object Segmentation and Detection algorithm. The second is the Feature Extraction algorithm.

The first sub-algorithm segments the needle tip, the Taylor cone, the jet and the deposited fiber `on the collector. In addition to that, the algorithm attempts to find the jet's deposition point on the collector. Finally, the segmented objects of interest are plotted for the user to visually inspect the output and assess the performance of the algorithm. To detect the objects of interest in each video frame we use the very much alike *meanshift* [31] and *camshift* [32] algorithms

The *meanshift* algorithm is based on a statistical concept directly related to clustering. Similar to other clustering algorithms, the *meanshift* algorithm scans the whole frame for high concentration of pixels of the same color. The main difference between the *meanshift* and the *camshift* algorithms is that the *camshift* algorithm has the capability to adjust, so that the tracking box can change its size and direction, to better correlate with the movements of the tracked object. The *meanshift* and *camshift* algorithm are useful tools to employ for object tracking. Also, unlike neural networks and other machine learning methods for object detection, these algorithms can be immediately implemented and deployed unsupervised, i.e. without the need to train a model with numerous labeled images. Instead, the algorithm takes as an input the initial color of the object, that needs to be detected, and then it tracks it throughout the rest of the video. On the other hand, using color as a primary method of identification, neither of the two algorithms can identify objects based on specific shapes and features, which makes them significantly less powerful than other methods. Furthermore, objects varying in color on a large scale and complex or noisy backgrounds can make object detection and tracking problematic. As a result, the *meanshift* and *camshift* algorithms work best under controlled environments.

The first step is to reverse the image colors so that the objects of interest are white and the background black. The next step is to apply a multi-color mask to segment them, and then to change the image color-space from BGR to HSV. Finally, the *meanshift* algorithm is applied to detect the needle and the Taylor cone, since there is barely any significant movement to them, as well as the *camshift* algorithm to detect and track the jet.

To find the deposition point, the algorithm needs to know the collector's position. Then, it creates a window around the collector, crops the region of interest from the frame and processes that instead of the whole frame. The built-in function used to find the deposition point is the *cv2.goodFeaturesToTrack*. This function finds the most prominent corner in our region of interest by calculating its eigen-values, as described in [33].

Finally, by subtracting the deposition point from the nozzle's position (center of blue rectangle in Figure 2c), we get the lag distance, which is depicted with a two-way orange arrow in **Figure 3**.

The second sub-algorithm is the one responsible for extracting all the jet features that are relevant to the process dynamics. These features are the diameter, areas, and angles of the jet as we move along the z-axis. Another important feature is the velocity of each jet's point along the x-axis relatively to the nozzle's position. To get all those features we follow a straightforward procedure. The algorithm takes three inputs, the first is the current video frame. The second input is the calibration factor ($cf$), which is a correlation between distance units (mm) and pixels. The last one is the stride. The stride indicates every how many pixels along the z-axis we perform computations. Using too small a stride would lead to more precise calculations but would tremendously increase the computation time. On the other hand, using too large a stride would lead to shorter computation times but at a risk to lose important information.

The first step is to change the frame's color-space from RGB scale to grayscale, so that the Canny edge detection algorithm[34] can be applied. The parameters of the Canny edge detector are [threshold_1, threshold_2] and were specified to 150 and 255 in a semi-automatic way, using trackbars while performing edge detection to other video samples. After performing Canny edge detection, we read the first row of pixels in our canny-frame, which now is an array of 0 and 255. If Canny algorithm has been implemented correctly when we read this row of pixels from left to right, the first time we encounter a 255 should be the left edge ($le$) of our jet. Likewise, the first time we encounter a 255 while reading the row of pixels from right to the left, should be the right edge ($re$) of our jet. By subtracting those two pixels' indices and multiplying with the calibration factor we get the diameter of the jet at this position in the z-axis, which is equal to 2 $R_j$.

$$2\,R_j = (re - le)cf \tag{1}$$

$$r_{boundary} = re \tag{2}$$

Those indexes are also stored in two variables ($le_{previous}, re_{previous}$) so that they can be used to calculate the jet angles as we move down the z-axis.

Then we repeat the procedure for every 'stride' rows. After finding the left ($le$) and right ($re$) edges and calculating the diameter, the area and angles can be calculated as:

$$A_j = \left[\left((re_{previous} - le_{previous}) + (re - le)\right) * stride\right] cf^2 \tag{3}$$

$$\theta_{jl} = \arctan((le - le_{previous})/stride) \tag{4}$$

$$\theta_{jr} = \arctan((re - re_{previous})/stride) \tag{5}$$

The $le_{previous}$, $re_{previous}$ are then updated with the $le$, $re$ values.

After accessing all frame's rows, the algorithm returns arrays containing all the quantified Diameters, Areas, Right Boundaries, Angles left and Angles right. The same procedure is applied to all frames. Right Boundaries are important because by subtracting the right edges of two consecutive frames we can calculate the jet's velocity ($U_j$) on the x-axis.

**6.2 Physics-based Modeling Module**

**Multiphysics Model.** The importance of accurately extracting jet properties is signified by several studies on predicting the jet stable region diameter, through mathematical modeling. Zhmayev et al. proposed a model by fully coupling the conservation of mass, momentum, charge and energy equations with a constitutive model and the electric field equations at the steady state [31]. Similar to most models, they utilize the thin filament approximation to obtain a simpler and more tractable solution. This assumption is possible by appropriately averaging the model variables across the radial direction. In addition, the charge and electric field equations are simplified, under the assumption of low electrical conductivity, as compared to the governing equations for isothermal simulations presented by Carroll and Joo [32]. The conservation of energy relation and a non-isothermal constitutive model were added to extend to non-isothermal situations. The resulting governing equations after being nondimensionalized are as follows (see Supplementary Information for Nomenclature):

**Continuity:**
$$R_j^2 V_j = 1 \tag{6}$$

**Momentum:**
$$Re V_j V_j' = Bo + \frac{\left(R_j^2(\tau_{zz} - \tau_{rr})\right)'}{R_j^2} + \frac{R_j'}{Ca R_j^2} + F_e \left[\sigma\sigma' + \beta_E E_t E_t' + \frac{2\sigma E_t}{R_j}\right] \tag{7}$$

**Charge:**
$$\sigma = R \tag{8}$$

**Electric field:**
$$E_t = \frac{1}{\left(1 + 2Z - \frac{Z^2}{x}\right)\left(\sqrt{1 + (R_j')^2}\right)} \tag{9}$$

$$E_t' = \frac{-2 + 2Z/x}{\left(1 + 2Z - \frac{Z^2}{x}\right)\left(\sqrt{1 + (R_j')^2}\right)}$$

**Energy:**
$$Pe V_j \Theta' = Na V_j'(\tau_{zz} - \tau_{rr}) - \frac{2 Bi_L (\Theta - \Theta_\infty)}{R_j} \tag{10}$$

**Constitutive:**
$$\tau_{zz} = \tau_{p,zz} + 2\beta f(\Theta) V_j' \tag{11}$$
$$\tau_{rr} = \tau_{p,rr} - \beta f(\Theta) V_j'$$
$$\tau_{p,zz} + \frac{De\Gamma}{\Theta + \Gamma}\left(\frac{\alpha \tau_{p,zz}^2}{1 - \beta} + f(\Theta)\left[V_j \tau_{p,zz}' - 2V_j' \tau_{p,zz} - \frac{V_j \tau_{p,zz} \Theta'}{\Theta + \Gamma}\right]\right) = 2(1 - \beta) f(\Theta) V_j'$$
$$\tau_{p,rr} + \frac{De\Gamma}{\Theta + \Gamma}\left(\frac{\alpha \tau_{p,rr}^2}{1 - \beta} + f(\Theta)[V_j \tau_{p,rr}' + V_j' \tau_{p,rr}]\right) = 2(1 - \beta) f(\Theta) V_j'$$
$$f(\Theta) = \exp\left[\frac{\Delta H}{R_{ig} \Delta T_{Rh}}\left(\frac{1}{\Theta + \Gamma} - \frac{1}{\Gamma}\right)\right]$$

The system of equations can be reduced to a set of five coupled first order ordinary differential equations (ODEs). Boundary Conditions are required, in order to proceed towards the numerical solution.

$$\tau_{p,zz}|_{Z=0} = 2(1 - \beta) f(\Theta) V_j' \tag{12}$$
$$\tau_{p,rr}|_{Z=0} = -(1 - \beta) f(\Theta) V_j' \tag{13}$$
$$\Theta|_{Z=0} = 0 \tag{14}$$
$$R|_{Z=0} = 1 \tag{15}$$
$$\left[\frac{6}{R_j^4}(R_j')^2 + \left(\frac{1}{Ca R_j^2} + Fe R_j\right) R_j' + \frac{2Fe}{\sqrt{1 + (R_j')^2}}\left(1 - \frac{\beta_E}{\sqrt{1 + (R_j')^2}}\right)\right]_{z=0} = 0 \tag{16}$$

**Table 2: Material properties of PCL**

| Property | Value |
|---|---|
| Zero-shear-rate (at 100°C) ($\eta_o$) | $1900\ Pa\ s$ |
| Relaxation time (at 100°C) ($\lambda_o$) | $0.019\ s$ |
| Activation energy of flow ($\Delta H/R_{ig}$) | $7938.4\ K$ |
| Density ($\rho$) | $1145\ kg/m^3$ |
| Heat capacity ($C_p$) | $1340\ J/kgK$ |
| Thermal conductivity ($k$) | $0.14\ W/mK$ |
| Electrical conductivity ($K$) | $9.5*10^{-9}\ S/m$ |
| Surface tension ($\gamma$) | $0435\ N/m$ |
| Ratio of solvent to zero-shear-rate viscosity ($\beta$) | 0.001 |
| Mobility factor ($\alpha$) | 0.015 |
| Dielectric constant ratio ($\varepsilon/\varepsilon_o$) | 2.9 |

**Table 3: Typical values of dimensionless parameters used for PCL**

| Parameter | Value |
|---|---|
| $Bi$ | 0.424 |
| $De$ | 1.14 |
| $Bo$ | 0 |
| $Re$ | $5.785*10^{-6}$ |
| $Ca$ | 1048.276 |
| $Na$ | 0.446 |
| $Fe$ | 0.0254 |
| $\Gamma$ | 21.283 |
| $Pe$ | 105.209 |
| $Pe_c$ | 0.1122 |

The model was implemented in Python. While true properties and parameters of the material are not provided the ones used in [13] for PCL were used. As also referred in [12], [13] the model slightly underpredicts the jet radius while in the Taylor cone area, but when the jet is stabilized, it accurately predicts it's radius. Knowing this, even if the volumetric flowrate (Q) is not provided with the dataset, a Particle Swarm Optimization (PSO) algorithm was also implemented to find the Q for which the predicted jet's radius better fits the computer vision observations.

**Geometrical Model.** Lag distance is a highly important parameter regarding the quality of the process outcome. Specifically, for some collector speeds, the jet falls onto the moving collector in a way reminiscent of a sewing machine, generating a rich variety of periodic patterns, such as meanders, W patterns, alternating loops and translated coiling (see **Figure 9d**). P. T. Brun et al. [33] proposed a quasistatic geometrical model, consisting of three coupled ordinary differential equations for the radial deflection, the orientation and the curvature of the path of the jet's contact point with the collector, capable of reconstructing the patterns observed experimentally while successfully calculated the bifurcation threshold of different patterns. They also evidenced that the jet/collector velocity ratio ($U_c/V_{jm}$) was the key factor for pattern variation.

According to this geometrical model, the deposited trace on the collector is a combination of the obit of the contact point (when collector's speed is equal to zero $U_c = 0$, the jet creates coiling patters with radius $R_c$) and the movement of the collector.

$$q(s,t) = r(s) + U_c \left(t - \frac{s}{V_{jm}}\right) e_x, \tag{17}$$

where $q(s,t)$ is the deposited trace, $s$ is the arc-length, $t$ is time, $r(s)$ is the contact point at time $s/V_{jm}$, $e_x$ is the direction of the collector's speed, $t - s/V_{jm}$ is the time that the contact point moves together with the collector. Differentiating $q(s,t)$ and moving from Cartesian to Polar coordinates ($r, \psi$ denote the polar coordinates of the contact point $r(s)$), and considering the curvature $\theta'$ at the bottom of the jet, we get the system of ODEs:

$$r' = \cos(\theta - \psi) + \frac{U_c}{V_{jm}} \cos\psi \tag{18}$$

$$\psi' = \frac{1}{r}\left(\sin(\theta - \psi) - \frac{U_c}{V_{jm}} \sin\psi\right) \tag{19}$$

$$\theta' = \frac{1}{R_c}\sqrt{\frac{r}{R_c}}\left(1 + \frac{0.715^2 \cos(\theta - \psi)}{1 - 0.715 \cos(\theta - \psi)} r\right) \sin(\theta - \psi) \tag{20}$$

This geometrical model was implemented in Python and by varying the dimensionless parameter $U_c/V_{jm}$ from 0 to 1 as suggested [30], the orbit and the deposited trace can be reconstructed. Verifying the results from [30], the critical velocity at which the straight pattern appears is $U_c = V_{jm}$, which means $U_c/V_{jm} = 1$. for speed ratios $0 < U_c/V_{jm} < 1$ the process is highly unstable, forming the translated coiling, alternating loops, W patterns and meanders when the speed ratios are 0.23, 0.48, 0.64, 0.83, respectively.

**6.3 Machine Learning Module**

**Gaussian Process Regression.** Gaussian Process Regression is a non-parametric stochastic process with strong probabilistic establishment [35]. GPR is a supervised machine learning technique, which predicts a probability distribution based on Bayesian theory unlike other machine learning algorithms that give deterministic predictions. The idea behind GPR is that the posterior probability can be modified based on a prior probability, given a new observation. Those characteristics allow the uncertainty quantification of each point prediction. Assuming there is a dataset available, consisting of input-output pairs of observations $D = \{x_i, y_i\} = (x, y)$, $i = 1, 2, \ldots, n$ that are generated by an unknown model function $f$

$$y = f(x), \quad x \epsilon \mathbb{R}^d \tag{21}$$

$f(x)$ can be completely estimated by a mean $m(x)$ and a covariance function $K(x, x')$.

$$m(x) = \mathbb{E}[f(x)] \tag{22}$$

$$K(x, x') = \mathbb{E}[(f(x) - m(x))(f(x') - m(x'))] \tag{23}$$

GPR aims to learn the mapping between the set of input variables and the unknown model $f(x)$, given the set of observations $D$. To map this correlation $f(x)$ is typically assigned a GP prior.

Gaussian Processes (GPs) are powerful modelling frameworks incorporating a variety of kernels. A Gaussian Process is a collection of random variables, any finite number of which have a joint Gaussian distribution [35].

$$f \sim \mathcal{GP}(m(x), k(x, x'; \theta)) \tag{24}$$

where $k$ is a kernel function with a set of trainable hyperparameters $\theta$. The kernel defines a symmetric-positive covariance matrix $K_{ij} = k(x_i, x_j; \theta)$, $K \in \mathbb{R}^{n \times n}$, which reflects the prior available knowledge on the function to be approximated. Furthermore, kernel's eigenvalues define a reproducing kernel Hilbert space, that determines the class of functions within approximation capacity of the predictive GP posterior mean. Hyper-parameters $\theta$ are trained by maximizing the marginal log-likelihood of the model [35].

Assuming a Gaussian likelihood and using the Sherman-Morrison-Woodbury formula the expression for the posterior distribution $p(f|y, X)$ is tractable and can be used to perform prediction given a new output $f_{n+1}$ for a new input $x_{n+1}$.

$$p(f_{n+1}|y_{1:n}, x_{1:n}, x_{n+1}) = \mathcal{N}(f_{n+1}|\mu_n(x_{n+1}), \sigma_n^2(x_{n+1})) \tag{26}$$

$$\mu_n(x_{n+1}) = k_{n+1} K^{-1} y_{1:n} \tag{27}$$

$$\sigma_n^2(x_{n+1}) = k(x_{n+1}, x_{n+1}) - k_{n+1} K^{-1} k_{n+1}^T \tag{28}$$

where $k_{n+1} = [k(x_{n+1}, x_1), \ldots, k(x_{n+1}, x_n)]$. As referenced before prediction consists of a mean, computed using the posterior mean $\mu_*$, and an uncertainty term, computed using the posterior variance $\sigma_*^2$.

**Multi-fidelity Modeling.** The GPR framework, presented above, can be extended to construct probabilistic models able to consider numerous information sources of different fidelity levels [24]. Supposing that $s$ levels of information source are available, the input, output data pairs can be organized by increasing fidelity as $D_t = \{x_t, y_t\}$, $t = 1, 2, \ldots, s$. So, $y_s$ denotes the output of the most accurate and expensive to evaluate model, while $y_1$ denotes the output of the cheapest and least accurate model to evaluate. Assuming that only two models are available, a high-fidelity model and a low fidelity model, the high-fidelity model can be defined as a scaled sum of the low fidelity model plus an error term:

$$f_{high}(x) = \rho f_{low}(x) + f_{err}(x) \tag{29}$$

where $\rho$ is a scaling constant quantifying the correlation between the two models and $f_{err}(x)$ denotes another GP which models the error.

A numerically efficient recursive inference scheme can then be constructed, by replacing the GP prior $f_{low}(x)$ with the GP posterior $f_{low_{n_{low}+1}}(x)$ of the previous inference level, while assuming that the corresponding experimental design sets $\{D_1, D_2, \ldots, D_s\}$ have a nested structure. This implies that the training inputs of higher fidelity model needs to be a subset of the training inputs of the low fidelity model. This scheme is matching totally the Gaussian posterior distribution predicted by the fully coupled scheme, only now the inference problem is decoupled into two GPR problems, yielding the multi-fidelity posterior distribution $p\left(f_{high}|y_{high}, X_{high}, f_{low_{n_{low}+1}}\right)$ with a predictive mean and variance at each level [18].

$$\mu_{low}(x_{n_{low}+1}) = \mu_{err} + k_{n_{low}+1} K_{low}^{-1} \left[y_{low_{1:n_{low}}} - \mu_{err}\right] \tag{30}$$

$$\mu_{high}(x_{n_{high}}) = \rho\mu_{low}\left(x_{n_{high}+1}\right) + \mu_{err} \tag{31}$$
$$+ k_{n_{high}+1} K_{high}^{-1}\left[y_{high_{1:n_{high}}} - \rho\mu_{low}\left(x_{high_{1:n_{high}}}\right) - \mu_{err}\right]$$
$$\sigma_{low}^2(x_{n_{low}+1}) = k(x_{n_{low}+1}, x_{n_{low}+1}) - k_{n_{low}+1} K_{low}^{-1} k_{n_{low}+1}^T \tag{32}$$
$$\sigma_{high}^2\left(x_{n_{high}+1}\right) \tag{33}$$
$$= \rho^2 \sigma_{low_{n_{low}+1}}^2\left(x_{n_{high}+1}\right) + k\left(x_{n_{high}+1}, x_{n_{high}+1}\right)$$
$$- k_{n_{high}+1} K_{high}^{-1} k_{n_{high}}^T$$

where $n_{high}$, $n_{low}$ denote the number of training points from the high and low fidelity models, respectively.

**Active Learning.** Let's assume again that $n$ observations are available $\{x_i, y_i\}, i = 1, \ldots, n$ where $y_i = f(x_i)$ and the next point to be evaluated $(x_{n+1}, y_{n+1})$ needs to be considered. The question that arises is if there is a more informed way to pick those points when evaluation is expensive to perform, rather than random picking.

This is achieved through an acquisition function $u(\cdot)$. The role of the acquisition function is to guide the search for the optimum. They are defined in a way such that high acquisition values correspond to a potential optimum of the unknown model $f$, large prediction uncertainty or a combination of those. Maximizing the acquisition function is used to select the next point to evaluate the function at. Consequently, the goal is to sample $f$ sequentially at $argmax_x u(x|D)$.

Every acquisition function depends on $\mu, \sigma^2$ or a combination of both. The scale at which it depends on each one of those defines the exploration-exploitation tradeoff. When exploring, points where the GP variance is large should be chosen. When exploiting, points where the GP mean is closest to the extremum should be chosen. Many acquisition functions are available, some of them are:

**Table 4: Types of acquisition functions for Active Learning scheme**

| | | |
|---|---|---|
| **Variance** | $\sigma^2(x)$ | Purely exploration, makes sure, that we learn the function $f$ everywhere on x to a similar level of absolute error. |
| **Probability of Improvement** | $PI(x) = \Phi\left(\dfrac{\mu(x) - f(x^+) - \xi}{\sigma(x)}\right),$ <br> $\Phi(\cdot)$ is the normal cumulative distribution function | Selects the point most likely to offer an improvement of at least $\xi$ but is extremely sensitive to the choice of the target. |
| **Expected Improvement** | $EI(x)$ $= \begin{cases}(\mu(x) - f(x^+) - \xi)\Phi(Z) + \sigma(x)\phi(Z), & \text{if } \sigma(x) > 0 \\ 0, & \text{if } \sigma(x) = 0\end{cases}$, <br> where $Z = \begin{cases}\frac{\mu(x) - f(x^+ - \xi)}{\sigma(x)}, & \text{if } \sigma(x) > 0 \\ 0, & \text{if } \sigma(x) = 0\end{cases}$, <br> $\phi(\cdot)$ is the normal probability distribution function | Similar to PI but takes into account the magnitude of the improvement a point can potentially yield as well |
| **Lower Confidence Bound** | $LCB(x) = \mu(x) - \kappa\sigma(x)$, $\kappa > 0$ | Selects points for evaluation based on the lower uncertainty bound |

After sampling $x_{n+1}$ and evaluating $f_{n+1}$, GP regression is performed to fit to the new point as well. Then the process repeats itself until termination criteria are met, such as a maximum number of iterations, a minimum or maximum value is reached, or uncertainty is below an allowed value.

# Supplementary Information

## 1. Nomenclature

**Parameters: Symbol – Description – Units**

| | Symbol | Description | Units |
|---|---|---|---|
| **Process** | | | |
| | $R_0$ | Needle tip radius | $mm$ |
| | $Z$ | Needle tip to collector distance | $mm$ |
| | $t$ | Time | $s$ |
| | $p$ | Pressure | $kPa$ |
| | $Q$ | Volumetric flowrate | $m^3/s$ |
| | $V_p$ | Voltage applied to needle tip | $kV$ |
| | $V_c$ | Voltage applied to collector | $kV$ |
| | $T_n$ | Needle temperature | °C |
| | $T_c$ | Collector temperature | °C |
| | $U_c$ | Collector speed | $mm\ s^{-1}$ |
| | $U_{cr}$ | Critical collector speed | $mm\ s^{-1}$ |
| **Jet** | | | |
| | $R$ | Jet radius | $mm$ |
| | $V_j$ | Jet speed in y axis | $mm\ s^{-1}$ |
| | $U_j$ | Jet speed in x axis | $mm\ s^{-1}$ |
| | $V_{jm}$ | Jet speed on impact point | $mm\ s^{-1}$ |
| | $\theta_{jl}$ | Angle left | (°) |
| | $\theta_{jr}$ | Angle right | (°) |
| | $A_j$ | Area | $mm^2$ |
| | $L_j$ | Lag distance | $mm$ |
| **Physics Jet Model** | | | |
| | $Re$ | Reynolds number | — |
| | $Pe$ | Peclet number | — |
| | $Ca$ | Capillary number | — |
| | $Fe$ | Electrostatic force parameter | — |
| | $Bo$ | Bond number | — |
| | $Na$ | Nahme-Griffith number | — |
| | $De$ | Deborah number | — |
| | $Bi_L$ | Local Biot number | — |
| | $\Theta$ | Dimensionless temperature | — |
| | $\Gamma$ | Temperature factor | — |
| | $\chi$ | Aspect ratio | — |
| | $\alpha$ | Mobility factor | — |
| | $\beta$ | Ratio of solvent to zero-shear-rate viscosity | — |
| | $\beta_E$ | Dielectric constant ratio | — |
| | $\sigma$ | Surface charge density | $C/m^2$ |

| | | |
|---|---|---|
| $E_T$ | The tangential component of the electric field to the jet surface | $V/m$ |
| $f(\Theta)$ | Temperature dependence of the zero-shear-rate viscosity | — |
| $\Delta H$ | Activation energy | $K$ |
| $R_{ig}$ | Ideal gas constant | — |
| $\Delta T_{Rh}$ | Temperature change necessary to substantially alter the rheological properties of the fluid | $K$ |
| $\tau_{zz}$ | Total axial normal stress | $Pa$ |
| $\tau_{rr}$ | Total radial normal stress | $Pa$ |
| $\tau_{p,zz}$ | Axial polymeric stress | $Pa$ |
| $\tau_{p,rr}$ | Radial polymeric stress | $Pa$ |
| **Geometrical model** | | |
| $R_c$ | Steady coiling radius | $mm$ |
| $\boldsymbol{q}$ | Jet's trace on the collector | $mm$ |
| $\boldsymbol{r}$ | The contact point | $mm$ |
| $s$ | Deposited jet's arc length | $mm$ |
| $r$ | Polar radius coordinate | $mm$ |
| $\psi$ | Polar angle coordinate | (°) |
| $\theta$ | Curvature at the bottom of the jet | (°) |
| **CV Metrology** | | |
| $t_{CV}$ | Processing time | $s$ |
| $fps$ | Frames per second | $s^{-1}$ |
| $cf$ | Calibration factor | $mm/pix$ |
| $stride$ | Indicating every how many pixels along the z-axis we perform computations | $pix$ |
| **Gaussian Processes** | | |
| $D$ | Dataset, available input-output pair of observation data | — |
| $f(\cdot)$ | Unknown function to be approximated | — |
| $m(\cdot)$ | Mean function determining the unknown function | — |
| $K(\cdot)$ | Covariance matrix determining the unknown function | — |
| $k(\cdot)$ | Kernel reflecting the prior available knowledge on the unknown function | — |
| $\theta$ | Kernels hyperparameters to be trained | — |
| $\mu(\cdot)$ | Mean prediction of GP model | — |
| $\sigma^2(\cdot)$ | Variance prediction of GP model | — |
| $n$ | Number of available data | — |
| **Multifidelity Modeling** | | |
| $f_{high}(\cdot)$ | High fidelity GP model | — |
| $f_{low}(\cdot)$ | Low fidelity GP model | — |
| $\rho$ | A scaling constant quantifying the correlation between the two models | – |
| $f_{err}(\cdot)$ | Another GP modeling the bias term for the high-fidelity data | |
| $n_{low}$ | Number of low-fidelity data available | – |
| $n_{high}$ | Number of high-fidelity data available | – |
| **Active Learning and Bayesian Optimization** | | |
| $u(\cdot)$ | Acquisition function | – |
| $\xi$ | Parameter specifying the least required improvement | – |
| $\Phi(\cdot)$ | The normal cumulative distribution function | – |
| $\phi(\cdot)$ | The normal probability distribution function | – |
| $\kappa$ | Parameter specifying reliability of confidence intervals | – |
| **Performance Metrics** | | |
| RMSE | The Root Mean Squared Error Performance Metric measures the average magnitude of the error between the predicted values and the true values. | - |
| MCIW | The Mean Confidence Interval Width Performance Metric measures the average width of a confidence interval. | - |
| Min. Regret | The Minimum Regret Performance Metric quantifies how well a method works at finding the optimum. | - |

## 2. Dataset

Video S1 and Video S2 published by Hrynevich et al.[10] were chosen as the source of the dataset used in this paper. A Sony Alpha 7 (Sony Corp. Japan) digital camera was used with a Nikon ED 200 mm lens (Nikon Corp. Japan). 1080 p resolution videos of the nozzle, jet and collector were taken at 50 frames per second. Process hyperparameters were set to 8 m s$^{-2}$ and 500 m s$^{-3}$ maximal stage acceleration and jerk, a 22G nozzle was used, polymer temperature was set to 87 °C and the voltage to the collector was set to -1.5kV, while the voltage to the nozzle was set to +5.75kV.

For Video S1 the air pressure feeding the nozzle was set to 1.2 bar and the distance between nozzle and collector was set to 3.5mm with a standard deviation 0.1mm. Collector's speeds tested in Video S1 were 191.25 mm s$^{-1}$, 212.5 mm s$^{-1}$, 255 mm s$^{-1}$, 340 mm s$^{-1}$, 510 mm s$^{-1}$, 850 mm s$^{-1}$, 1530 mm s$^{-1}$ and 2890 mm s$^{-1}$.

For Video S2 the air pressure feeding the nozzle was set to 2.4 bar and the distance between nozzle and collector was set to 4.5mm with a standard deviation 0.1mm. Collector's speeds tested in Video S2 were 292.5 mm s$^{-1}$, 520 mm s$^{-1}$, 1300 mm s$^{-1}$ and 4420 mm s$^{-1}$.

First, the videos were split based on the collector speed setting. Second, the video frames were cropped to remove redundant pixels that would result to increased processing time. For real time video processing the user would need to specify the region of interest in the frame, so as to crop and dispose needless information, as well as the position of the nozzle, the collector, and a factor, which represents the length of the Taylor cone depending on the nozzle's diameter.

## 3. Supporting Figures

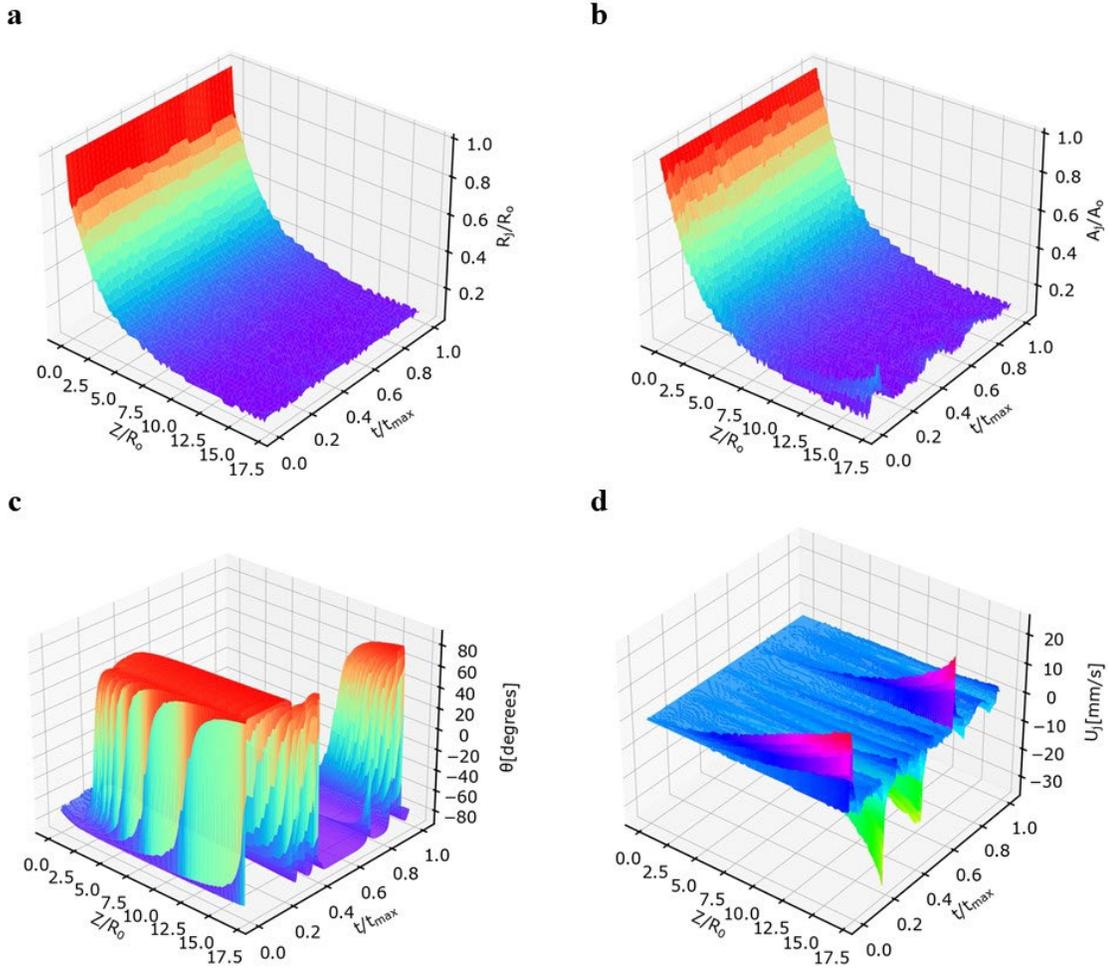

**Figure 1: Features Extracted from Computer Vision Module.** **a)** Normalized jet radius ($R_j/R_o$) obtained from the computer vision metrology module of the GPJet framework plotted against the normalized jet length ($Z/R_o$) and the normalized time ($t/t_{max}$). **b)** Normalized jet area ($A_j/A_o$) obtained from the computer vision metrology module of the GPJet framework plotted against the normalized jet length ($Z/R_o$) and the normalized time ($t/t_{max}$). **c)** Jet angles ($\theta$) obtained from the computer vision metrology module of the GPJet framework plotted against the normalized jet length ($Z/R_o$) and the normalized time ($t/t_{max}$). **d)** Jet velocities ($U_j$) obtained from the computer vision metrology module of the GPJet framework plotted against the normalized jet length ($Z/R_o$) and the normalized time ($t/t_{max}$).

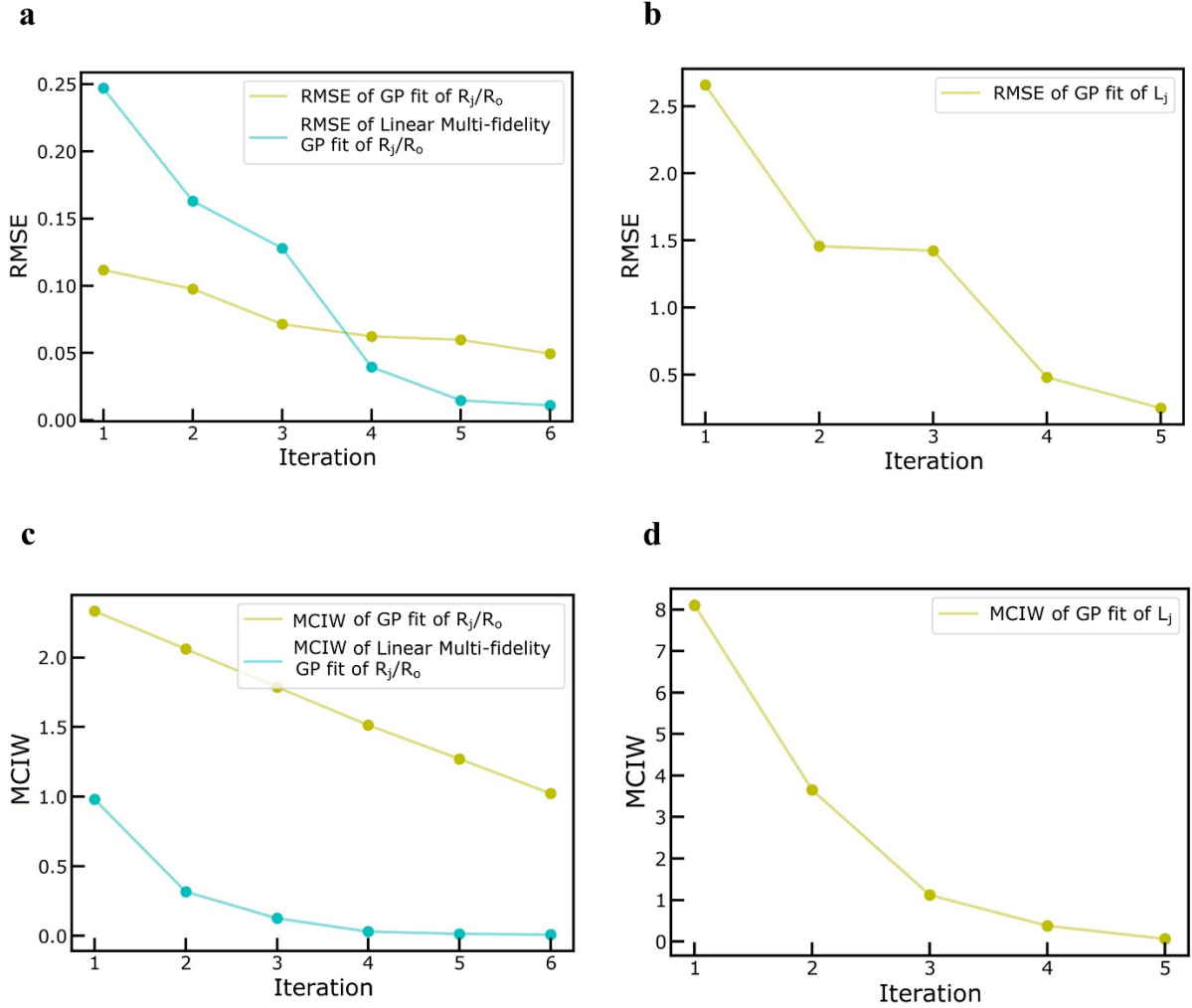

**Figure 2: Performance Metrics Evolution for Active Learning Tasks. a)** Root Mean Squared Error (RMSE) evolution after each iteration, regarding the normalized jet radius ($R_j/R_o$). **b)** Root Mean Squared Error (RMSE) evolution after each iteration, regarding the lag distance ($L_j$). **c)** Mean Confidence Interval Width (MCIW) evolution after each iteration, regarding the normalized jet radius ($R_j/R_o$). **d)** Mean Confidence Interval Width (MCIW) evolution after each iteration, regarding the lag distance ($L_j$).

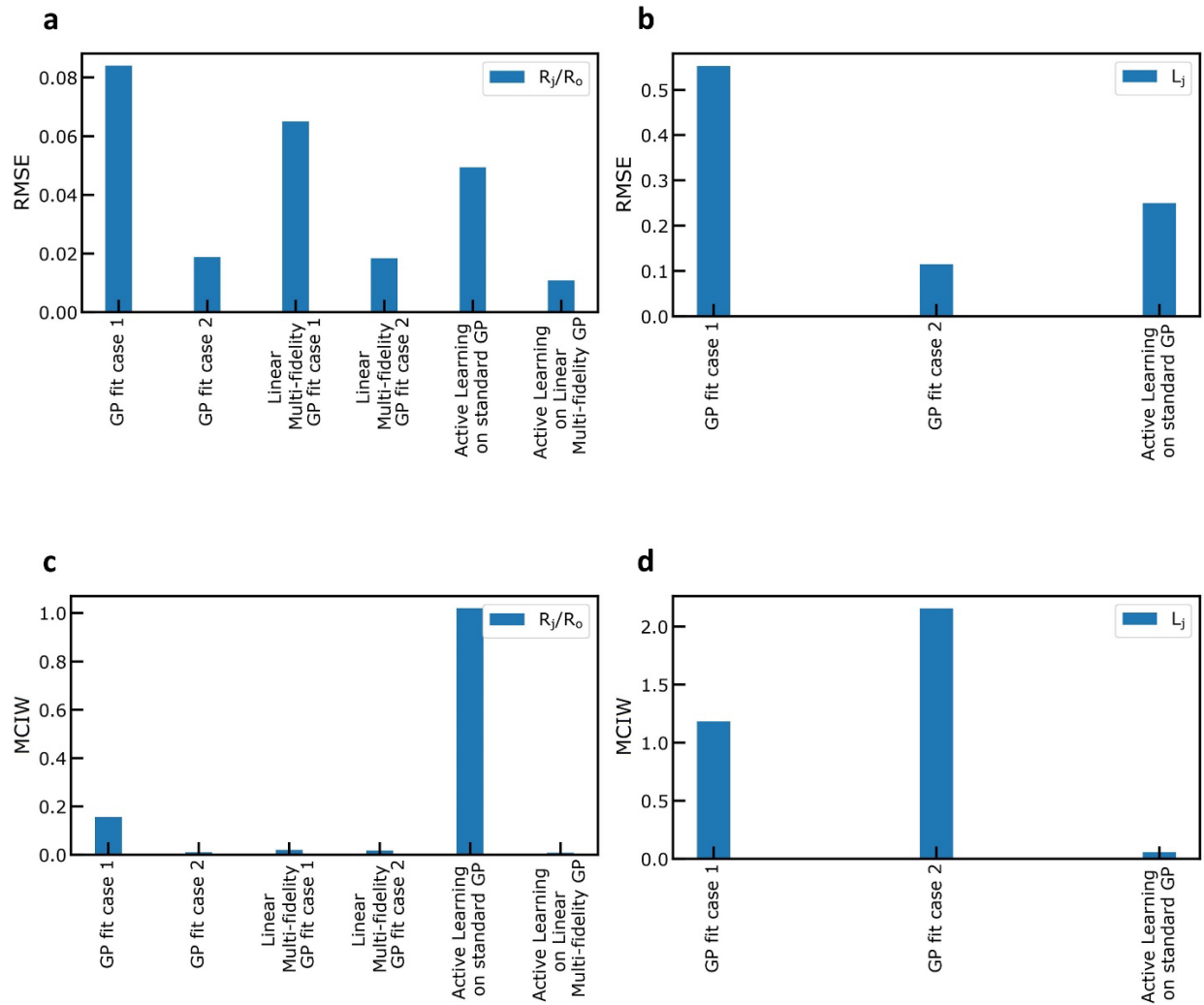

**Figure 3: Collective Performance Metrics for Regression and Active Learning Tasks. a)** Root Mean Squared Error (RMSE) for every case, regarding the normalized jet radius ($R_j/R_o$). **b)** Root Mean Squared Error (RMSE) performance metric, for every case, regarding the lag distance $L_j$. **c)** Mean Confidence Interval Width (MCIW) performance metric, for every case, regarding the normalized jet radius ($R_j/R_o$). **d)** Mean Confidence Interval Width (MCIW) performance metric, for every case, regarding the lag distance ($L_j$).

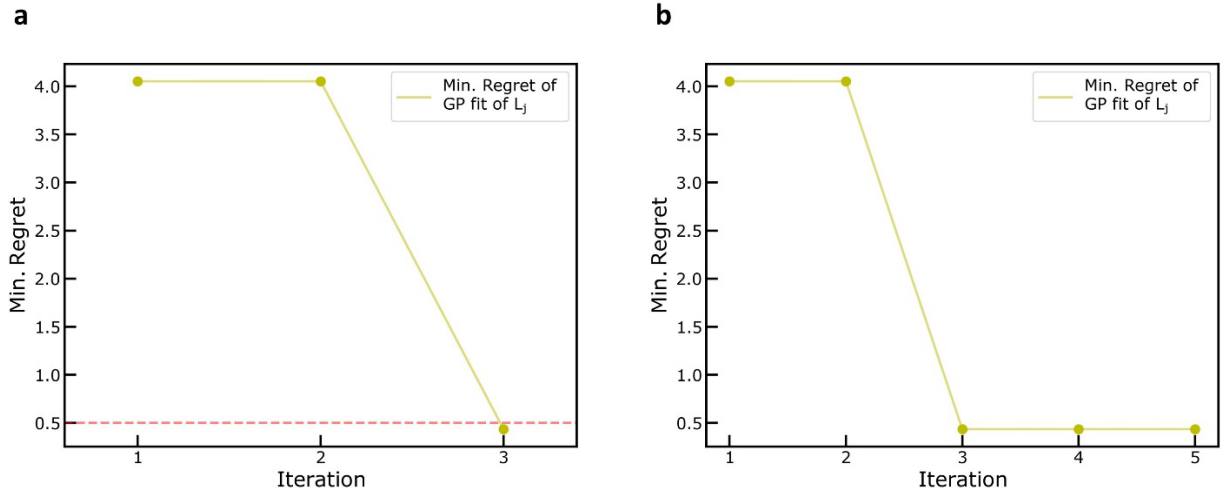

**Figure 4: Performance Metrics. a)** Minimum Regret Performance Metric evolution, after each iteration, regarding the Bayesian Optimization Task to find the minimum lag-distance ($L_j$). **b)** Minimum Regret Performance Metric evolution, after each iteration, regarding the Active Learning Task to explore the design space of lag-distance ($L_j$) for specific speed ratios ($U_c/V_{jm}$).